\documentclass[letterpaper, 10 pt, conference]{ieeeconf}  % Comment this line out if you need a4paper

\IEEEoverridecommandlockouts                              % This command is only needed if 
\usepackage{cite}
\usepackage{amsfonts}

\usepackage{siunitx}

\usepackage{microtype}
\usepackage{graphicx}
\usepackage{subfigure}
\usepackage{booktabs} % for professional tables

\usepackage{hyperref}
\newcommand{\beq}{\begin{equation}}
\newcommand{\eeq}{\end{equation}}
\def\be {\begin{equation}}
\def\ee {\end{equation}}
\def\bs#1\es{\begin{split}#1\end{split}}
\def\ba#1\ea{\begin{align}#1\end{align}}
\def\baed#1\eaed{\begin{aligned}#1\end{aligned}}
\def\bged#1\eged{\begin{gathered}#1\end{gathered}}
\def\bea{\begin{eqnarray}}
\def\eea{\end{eqnarray}}

\usepackage{amsmath}

\usepackage{hyperref}
\hypersetup{
colorlinks=true
,urlcolor=blue
,anchorcolor=blue
,citecolor=blue
,filecolor=blue
,linkcolor=blue
,menucolor=blue
,pagecolor=blue
,linktocpage=true
,pdfproducer=medialab
,pdfa=true
}

\title{\LARGE \bf
A Quadratic Actor Network for Model-Free Reinforcement Learning
}

\author{Matthias Weissenbacher$^{1}$ and Yoshinobu Kawahara$^{1,2}$% <-this % stops a space

\thanks{$^{1}$ RIKEN Center for Advanced Intelligence Project, Japan}%
\thanks{$^{2}$ Institute of Mathematics for Industry, Kyushu University, Japan}%
\thanks{*Equal contribution}% <-this % stops a space
}

\begin{document}

\maketitle
\thispagestyle{empty}
\pagestyle{empty}

%%%%%%%%%%%%%%%%%%%%%%%%%%%%%%%%%%%%%%%%%%%%%%%%%%%%%%%%%%%%%%%%%%%%%%%%%%%%%%%%

\begin{abstract}

In this work we discuss the incorporation of quadratic neurons into policy networks in the context of model-free actor-critic reinforcement learning.
Quadratic neurons admit an explicit quadratic function approximation in contrast to conventional approaches where the the non-linearity is induced by the activation functions. 
We perform empiric experiments on several MuJoCo continuous control tasks and find that when quadratic neurons are added to MLP policy networks  those outperform the baseline MLP whilst admitting a smaller number of parameters. The top returned reward is  in average increased by $5.8\%$ while  being  about $21\%$ more sample efficient.  Moreover, it can maintain its advantage against added action and observation noise.

%In particular the latter property is of interest for a Sim2Real transfer. 

\end{abstract}

%%%%%%%%%%%%%%%%%%%%%%%%%%%%%%%%%%%%%%%%%%%%%%%%%%%%%%%%%%%%%%%%%%
%%%%%%%%%%%%%%%%%%%%%%%%%%%%%%%%%%%%%%%%%%%%%%%%%%%%%%%%%%%%%%%%%%
\section{INTRODUCTION}
\label{Introduction}
%%%%%%%%%%%%%%%%%%%%%%%%%%%%%%%%%%%%%%%%%%%%%%%%%%%%%%%%%%%%%%%%%%

\begin{figure}[ht]
\vskip 0.2in
\vspace{-0,2 cm}
\begin{center}
\centerline{\includegraphics[width=\columnwidth]{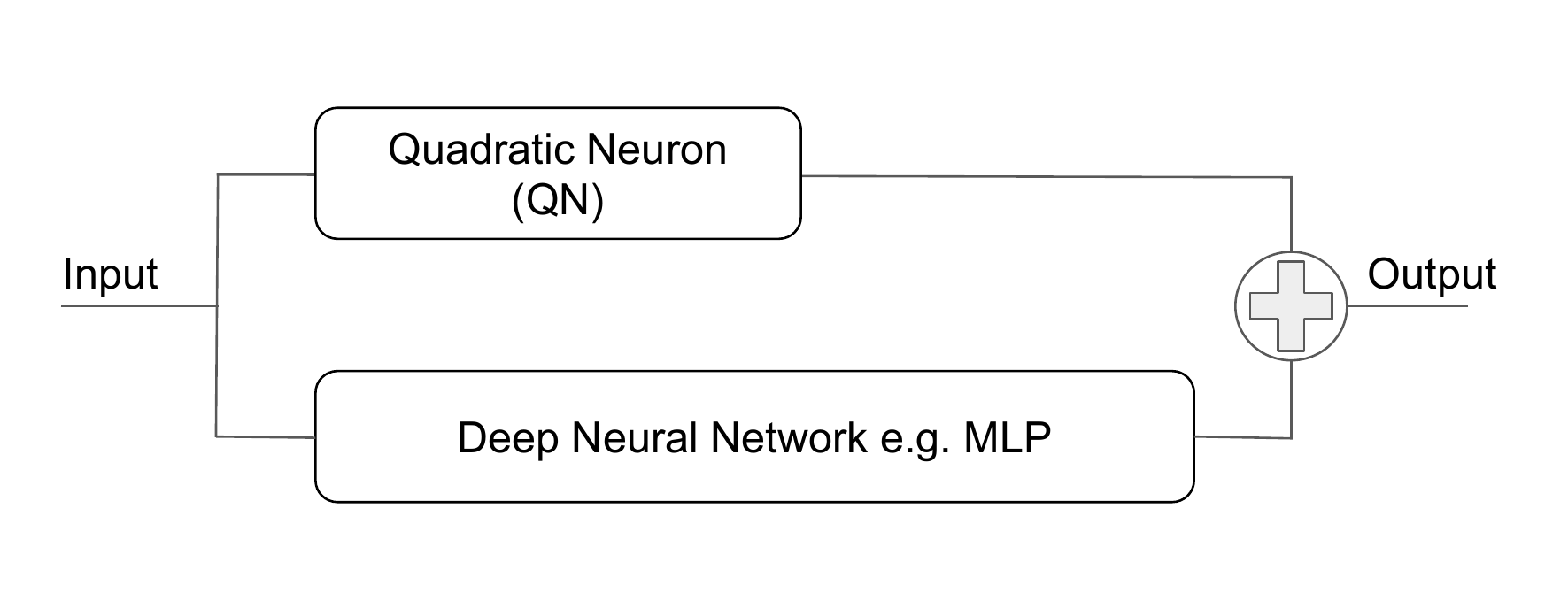}}
\vspace{-0,35 cm}
\caption{Quadratic neuron connecting input and output in a single layer additively to  a deep neural network architecture in our study given by a MLP. }
\label{QN}
\end{center}
\vskip -0.2in
\end{figure}

Underlying all reinforcement learning (RL) algorithms is the simple mechanism of  receiving a reward feedback  upon an interaction with the environment.  This allows the algorithms to acquire sophisticated skills. Deep reinforcement learning has thus been successfully applied to a variety of tasks such as robotics \cite{HaarnojaPZDAL18} as well as  board and video games \cite{Silver,Vinyals,samvelyan2019starcraft}. In this work we focus on model-free algorithms which can be readily applied to various environments. In particular,  online Actor-Critic methods for challenging settings with high-dimensional action and state spaces have  recently gained a lot of attention  \cite{mnih2016asynchronous,lillicrap2019continuous,haarnoja2018soft,fujimoto2018addressing}. The common perspective is that their success is owed to the use of  deep neural networks which allow for  high-capacity function approximations. The default policy network for the Actor is usually a Multi-Layer-Perceptron (MLP).
The majority of the  recent literature  - with a few notable exceptions \cite{srouji2018structured,liu2019recurrent} - attempts to improve learning results by altering the algorithm e.g. by modifying the loss function and interaction of the Actor and Critic. In contrast our angle is to study other policy networks with the aim of improving the general reward performance while reducing the number of trainable weights and being more computationally resource efficient. We achieve this by introducing a  single layer quadratic neuron in between input and output in addition to the deep MLP in a network called  Q-MLP (Quadratic-MLP), see fig.\,\ref{QN}.
The non-linearity of a conventional MLP is induced by the activation functions. In contrast, the Q-MLP allows for a direct approximation using the quadratic dependencies of the input as a first order optimisation problem.
Thus naturally using  quadratic layers one gains expressive advantages \cite{FAN2020383}.
We introduce the Q-MLP only for the Actor network while we do not alter the Critic network i.e.\,we use a conventional MLP for the latter.
For our empirical tests we focus on the two following  state-of-the-art model-free algorithms: SAC \cite{pmlr-v70-haarnoja17a} and TD3 \cite{fujimoto2018addressing}. Our source code as well as the raw data for our main results can be found online\footnote{\url{https://github.com/matthias-weissenbacher/Quadratic_MLPs_in_RL}}. 
Let us emphasize that we do not modify the hyperparamters of the TD3 and SAC algorithm to accommodate for the Q-MLP actor policy. Regardless, we find a benefit across almost all tasks. The top returned reward for the TD3  is  in average increased by $8.9\%$ while  being  about $29.98\%$ more sample efficient. The found improvements for the SAC algorithm are more moderate: $2.7\%$  and $12.3\%$ for  top returned reward and sample efficiency, respectively. 
These numbers can be improved when relaxing the condition of reducing the number of weights of the Q-MLP in comparison to the MLP. E.g.\,we show that in the case of the Ant-v3 environment the benefit doubles from $\sim 16 \%$ to $\sim 33 \%$ for the top reward and from $\sim 40\%$ to $\sim 60\%$ for the sample efficiency compared to the MLP when increasing the number of weights of the quadratic layer.
Moreover, it is worth noting that the benefit over the conventional MLP is maintained when action or observation noise is added.\footnote{Experiments for a smaller number of environments.} For specific setups we even improve the noise robustness compared to the MLP. Noise robustness  of deep neural nets is crucial for Sim2Real transfer, see e.g.\cite{zhao2020simtoreal} for a review. In particular, let us emphasise that our method does not require domain randomisation \cite{Fernndez2020DeepRL,Kaspar2020Sim2RealTF}.

Let us comment on another angle under which one may view our proposal. Reduction techniques of fully-connected neural networks to sparser representations with similar output performance have been widely investigated \cite{LeCun90,NIPS1992_647,han2015learning,hinton2015distilling,frankle2018lottery}.  Those are mostly  found in supervised learning research. The lottery ticket hypothesis states that “dense,  randomly-initialized, feed-forward networks contain subnetworks that … reach test accuracy comparable to the original network”\cite{frankle2018lottery}. 
The degrees of freedom of  function approximation of the the Q-MLP are contained in a more parameter extensive MLP  which naturally incorporates quadratic dependencies over the activation functions.  
Thus one may view the Q-MLP as a "sparser version" of the latter. One may thus consider our results as a step forward towards identifying a lottery ticket winner of neural networks in reinforcement learning.

The success of the Q-MLP may be attributed to the fact that it allows to propagate quadratic dynamic dependencies directly from input to output. In other words,  the Q-MLP Actor Policy can thus adapt easily to gain potential benefits in exploring these quadratic directions. This seems beneficial in particular in an online reinforcement setting where dynamic exploration is the source for new data. 
Although not discussed explicitly  in this work it is expected that our approach may as well lead to benefits in an offline reinforcement learning.

%%%%%%%%%%%%%%%%%%%%%%%%%%%%%%%%%%%%%%%%%%%%%%%%%%%%%%%%%%%%%%%%%%
%%%%%%%%%%%%%%%%%%%%%%%%%%%%%%%%%%%%%%%%%%%%%%%%%%%%%%%%%%%%%%%%%%
\section{RELATED WORK}
\label{Related_work}
To the best of our knowledge this work presents the first study of a quadratic layer in combination with a MLP  as the  policy network in the context of reinforcement learning. 
The standard policy network commonly used for the Actor in continuous control tasks is the Multi-Layer Perceptron (MLP). The non-linearity of deep neural networks as in the MLP is introduced by activation functions. Recently quadratic layers have been used in the context of categorical image classification in supervised learning \cite{fan2017new,CNNq2,fan2019universal,xu2020efficient}.
We incorporate the quadratic neuron as a single unit between input and output additively to a MLP which is in contrast to \cite{xu2020efficient}. \footnote{Another distinguishing minor feature contrasting our work from \cite{xu2020efficient} is that we make us of activation functions in the final layer. } Moreover, their study is restricted to supervised learning setups.

\begin{figure}[ht]
\vspace{-0,7 cm}
\vskip 0.2in
\begin{center}
\centerline{\includegraphics[width=\columnwidth]{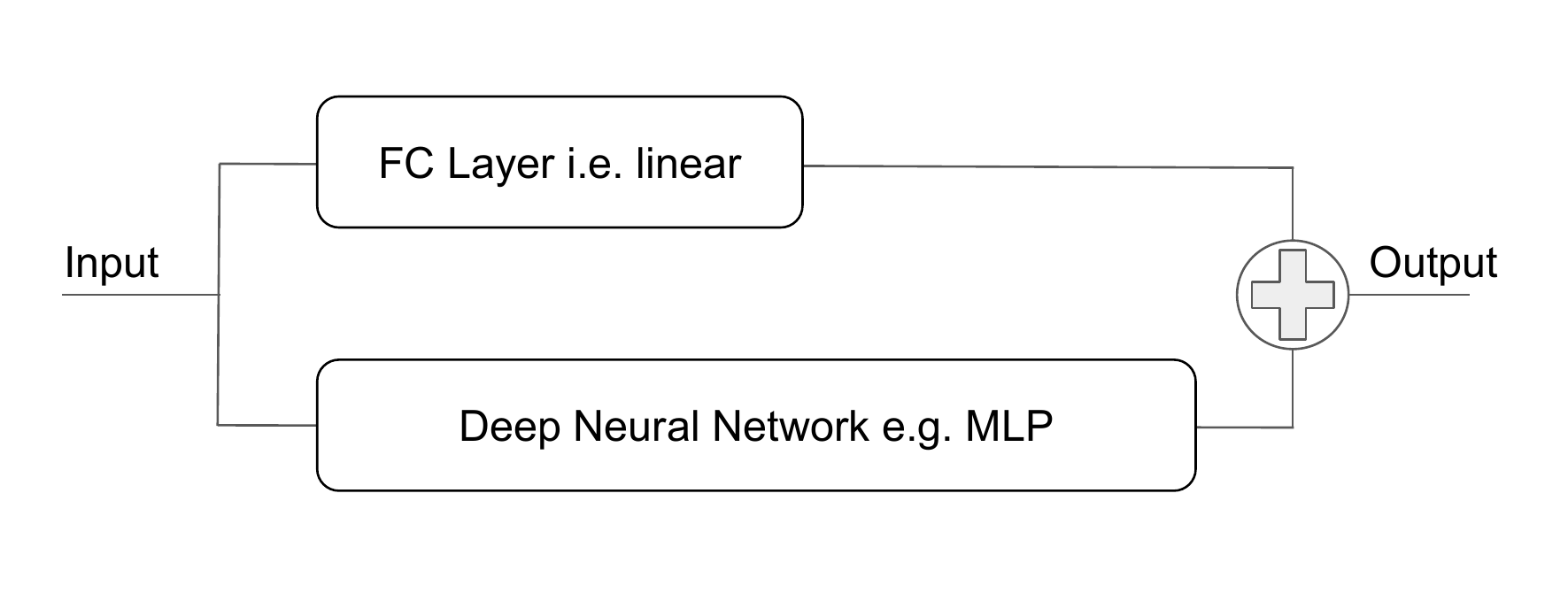}}
\vspace{-0,5cm}
\caption{Linear neuron  i.e. a fully connected (FC) layer connecting input and output in a single layer additively to a deep neural network architecture e.g. a MLP. }
\label{L-MLP}
\end{center}
\vskip -0.2in
\end{figure}

\begin{figure}[ht]
\vspace{-0,7 cm}
\vskip 0.2in
\begin{center}
\centerline{\includegraphics[width=\columnwidth]{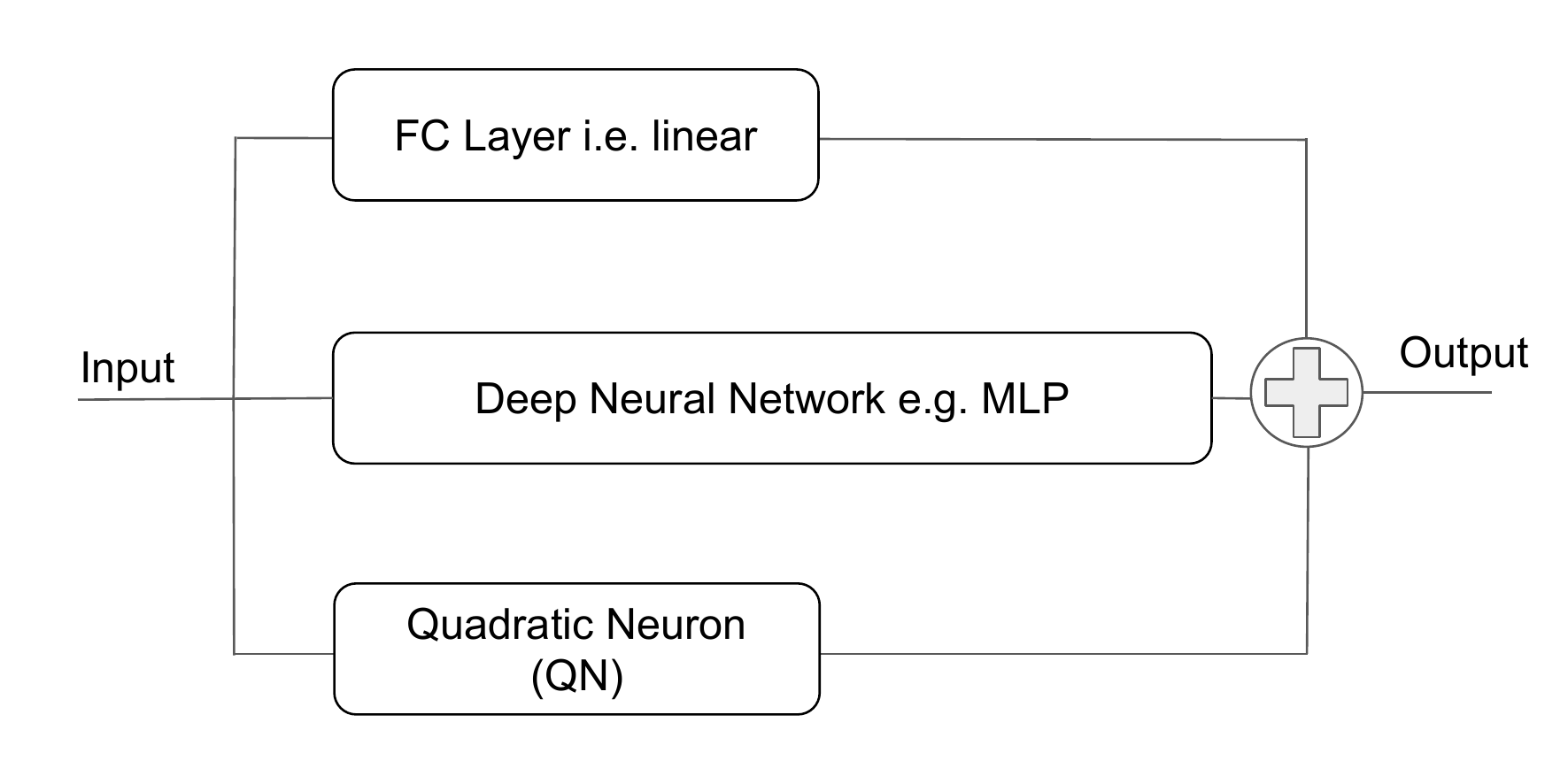}}
\vspace{-0,3 cm}
\caption{Linear and quadratic neuron   connecting input and output each in a single layer additively to a deep neural network architecture  e.g. a MLP. }
\label{LQ-MLP}
\end{center}
\vskip -0.2in
\end{figure}

More relevant for us is the previously discussed  architecture  \cite{he2015deep,srouji2018structured}  employed in the context of reinforcement learning. Their approach is conceptually analog to ours  as their architecture is comprised of a linear module which directly connects the input state additively to the output state of a deep neural network i.e. the MLP, see fig.\,\ref{L-MLP}. The authors refer to this architecture as "structured control nets". In other words, the authors split the policy network into linear and nonlinear components where the latter is given by a MLP. 
 This is in contrast to our approach where the quadratic neuron  layer  does not admit any linear components but is of quadratic order instead, see fig.\,\ref{QN}.  Moreover, it is worth emphasizing that as the quadratic neuron  uses a quadratic function approximation it carries more complexity. Lastly, we allow for activation functions i.e. $tanh$ to act upon the sum of the quadratic neuron and the MLP output.
Due to the similarity to \cite{srouji2018structured} we provide results on their architecture in the ablation study, see fig.\,\ref{L-MLP}. Moreover, we provide the ablation study for the case where a quadratic neuron and FC layer are present, see fig.\,\ref{LQ-MLP}.
Lastly, let us mention the idea of recurrent strategies in RL dates back two decades \cite{LSTM_RL}. More recent advances build on the above strategy \cite{liu2019recurrent,LeCun90}. Our results give comparable benefits while being supported by a much simpler architecture.\footnote{As our focus is on simple Actor  policy architectures we do not consider LSTMs in the main scope of this work. And thus do not perform an empirical quantitative comparison of the Q-MLP with an LSTM architecture.}

%%%%%%%%%%%%%%%%%%%%%%%%%%%%%%%%%%%%%%%%%%%%%%%%%%%%%%%%%%%%%%%%%%

%%%%%%%%%%%%%%%%%%%%%%%%%%%%%%%%%%%%%%%%%%%%%%%%%%%%%%%%%%%%%%%%%%
%%%%%%%%%%%%%%%%%%%%%%%%%%%%%%%%%%%%%%%%%%%%%%%%%%%%%%%%%%%%%%%%%%
\section{BACKGROUND}
\label{Background}

%%%%%%%%%%%%%%%%%%%%%%%%%%%%%%%%%%%%%%%%%%%%%%%%%%%%%%%%%%%%%%%%%%

In this work we focus on two recent state-of-the-art model-free algorithms SAC \cite{pmlr-v70-haarnoja17a} and TD3 \cite{fujimoto2018addressing}. SAC is an algorithm for soft Q-learning \cite{Rawlik,pmlr-v70-haarnoja17a, Fox2016TamingTN} which estimate a “soft Q-function” directly. The latter implicitly induces a maximum entropy policy. 
On the other hand the TD3 algorithm is the direct successor of DDPG \cite{lillicrap2019continuous, pmlr-v32-silver14}. It owes it improvement over DDPG  to the following three major changes: clipped double Q-Learning and delayed policy as well as target policy smoothing.

Let us remark that as a future extension of our present  work it would be interesting to incorporate Q-MLP into other types of algorithms for scalable learning for continuous control problems such as Trust-Region Policy Optimisation (TRPO)\cite{schulman2017trust} and its derivative class of Proximal Policy Optimisation algorithms  (PPO) \cite{schulman2017proximal} as well as Maximum a Posteriori Policy Optimisation (MPO) \cite{abdolmaleki2018maximum,pmlr-v119-abdolmaleki20a}. However, such a study is beyond the scope of this work.

\subsection{Preliminaries}

Reinforcement learning algorithms train policies to maximize the cumulative reward received by an agent who interacts with an environment. The underlying setting is given by a Markov decision process $ (\mathcal{S}, \mathcal{A}, p, r, \gamma)$, with S  being a set of states,  $\mathcal{A}$  the action space and  p  the transition density function. Moreover,  $\gamma$ is the discount factor and r the reward function. 
At any discrete time the agent chooses an action $a_t \in \mathcal{A}$ according to its underlying  policy $\pi_\theta (a_t | s_t) $ based on the information of the current state $s_t \in \mathcal{A}$. The policy is parametrized by $\theta$.
While the policy of the TD3 is deterministic  i.e. it predicts a single action with probability one,  the policy of the SAC algorithm is of stochastic nature. The parameters $\theta$ naturally are the weights in the neural network function approximation of the Actor as well as the Critic.
The agent i.e. the Actor-Critic is trained  to maximize the expected $\gamma$-discounted cumulative reward
\begin{equation}
\mathcal{J}_{\pi}(\theta) =  {\bf E}_{\pi} \Big[ \sum_{t=0}^T \gamma^ t \, r_\pi(s_t,a_t) \Big] \;\; , 
\end{equation}
with respect to the policy network parameters  $\theta$.

%%%%%%%%%%%%%%%%%%%%%%%%%%%%%%%%%%%%%%%%%%%%%%%%%%%%%%%%%%%%%%%%%%
%%%%%%%%%%%%%%%%%%%%%%%%%%%%%%%%%%%%%%%%%%%%%%%%%%%%%%%%%%%%%%%%%%
\section{Q-MLP — Quadratic-Multi-Layer-Perceptron}
\label{Q-MLP}
%%%%%%%%%%%%%%%%%%%%%%%%%%%%%%%%%%%%%%%%%%%%%%%%%%%%%%%%%%%%%%%%%%
A quadratic neuron is a deep neural network layer that outputs a vector of quadratic monomials \cite{book_QN_mention} constructed from the input e.g. for the input $ {\bf{x}} =(x_1,x_2,x_3)$  the output would be  $ y = \Theta_{11} x_1^ 2 + \Theta_{12} x_1 x_2, +\Theta_{13} x_1 x_3+ \Theta_{22} x_2^ 2+\Theta_{23} x_2 x_3+\Theta_{33} x_3^ 2$ where the $\Theta$'s are six trainable weights. The latter can be arranged in an upper triangular weights matrix $\hat \Theta$ which allows one to generally express the action of the quadratic neuron  as
\beq\label{QNdef}
y^ f = \sum_{\stackrel{j=1}{ j \geq i}}^ N \sum_{i=1}^ N   \hat\Theta^ f_{ij} x_i x_j= {\bf{x}}^T\, \cdot \, \hat \Theta^f \, \cdot {\bf{x}} \;.
\eeq
Where $f=1,\dots, n_f$ with $n_f$ the number  of features of the quadratic neuron and $\bf x$ a N-dimensional input vector.
The trainable degrees of freedom of a quadratic neuron are given by the matrix elements of $\hat\Theta^f$ i.e. its upper triangular components which are counted as 
\beq\label{QN_original}
\tfrac{1}{2} N \cdot (N +1) \cdot n_f \;\;.
\eeq 
One notes the $\mathcal{O}(N^ 2)$ scaling which makes an efficient direct implementation of eq.\,\eqref{QNdef} difficult for large hidden feature sizes. As proposed recently \cite{xu2020efficient} one may simply take the common "square" of a single linear i.e. fully connected layer to approximate the degrees of freedom of eq. \eqref{QNdef}.  This reduces the scaling to $\mathcal{O}(N)$  as well utilizes the parallelisation benefits build into deep learning libraries \cite{NEURIPS2019_9015, tensorflow2015-whitepaper}.  Thus we   adopt the following implementation for the quadratic neuron  in our work
\beq\label{QNsimpledef}
y_ f := \mathcal{Q}_f(\bf x) = \Big( \sum_{i=1}^ N   \hat\Theta'_{f i} x_{i}\Big) *\Big( \sum_{j=1}^ N   \hat\Theta''_{f j} x_j \Big)\;.
\eeq
where $\hat\Theta' $ and $ \hat\Theta''$ are weight matrices. Note that the index $f$ appears in both brackets in eq.\,\eqref{QNsimpledef} which denotes that the multiplication $*$ is element-wise.  Moreover, let us emphasize that the quadratic neuron in our neural network serves as a direct connection between input and output, see figure \ref{QN}. Thus the implementation \eqref{QNsimpledef} is  too restrictive as the input dimension is equal to the environments state vector and the feature dimensions equal to the one of the environments action state. Thus to allow for a higher number of trainable weights we choose the intermediate features size larger  than the environments action state and introduce another linear layer as
\beq\label{QNsimpledef2}
y_ a := \sum_{f=1}^ {n_f :=\text{hidden features}} \hat\Theta_{af} \mathcal{Q}_f (\bf x) + b_a \;\; ,
\eeq
where $\hat\Theta , {\bf b}$ are the weight matrix and bias vector, respectively, while $a=1,\dots, $ number of action dimension of the environment.  Eq. \eqref{QNsimpledef2}  simply represents  another fully connected layer acting on \eqref{QNsimpledef}. The crucial distinction to a MLP is that no activation function is used which would introduce additional non-linearities. Concludingly, \eqref{QNsimpledef2} is of precisely quadratic order in the monomials of the input vector.

Our guiding principle is to reduce the number of weights compared to the three layer actor policy MLP network from \cite{wei2021fork} i.e. two hidden layers with hidden unit dimension $256$ each. For the deep neural network part of our Q-MLP we choose the same three layer MLP architecture each with two hidden unit dimensions either $n_h = 64,128,192$ .  For the features for the quadratic neuron  we are guided by the relation \eqref{QN_original}  to arrive at
\beq \label{featureFormula}
n_f = \text{integer-part}\big(\tfrac{\kappa^ 2}{2}  \cdot   N  \cdot (N+1)\big) \;\;, \;\; \text{with} \;\; 0 < \kappa \leq 1 \;\; ,
\eeq 
where $N$ is the input  dimensions i.e. the observation space dimension of the environment. The expression \eqref{featureFormula} provides a convenient guiding principle for choosing the hyper-parameter  $n_f$ which served us well in achieving good empirical results without almost any hyper-parameter search.
%%%%%%%%%%%%%%%%%%%%%%%%%%%%%%%%%%%%%%%%%%%%%%%%%%%%%%%%%%%%%%%%%%
%\subsection{Formal defintion}
%\label{Formal_defintion}
%%%%%%%%%%%%%%%%%%%%%%%%%%%%%%%%%%%%%%%%%%%%%%%%%%%%%%%%%%%%%%%%%%
 
%%%%%%%%%%%%%%%%%%%%%%%%%%%%%%%%%%%%%%%%%%%%%%%%%%%%%%%%%%%%%%%%%%
\section{EXPERIMENTS}
\label{Experiments}

This section contains the empirical tests of our Q-MLP actor policy compared to the MLP baseline  in \ref{sec:mainResults} as well as an ablation study in subsection \ref{sec:ablation}.

\subsection{Training Environments \& RL Algorithm }
\label{sec:mainResults}
\begin{figure}
\vspace{0.20 cm}
   \centering
   \tabcolsep =  0.5pt
\begin{tabular}{ccc}
\includegraphics[width=2.8cm]{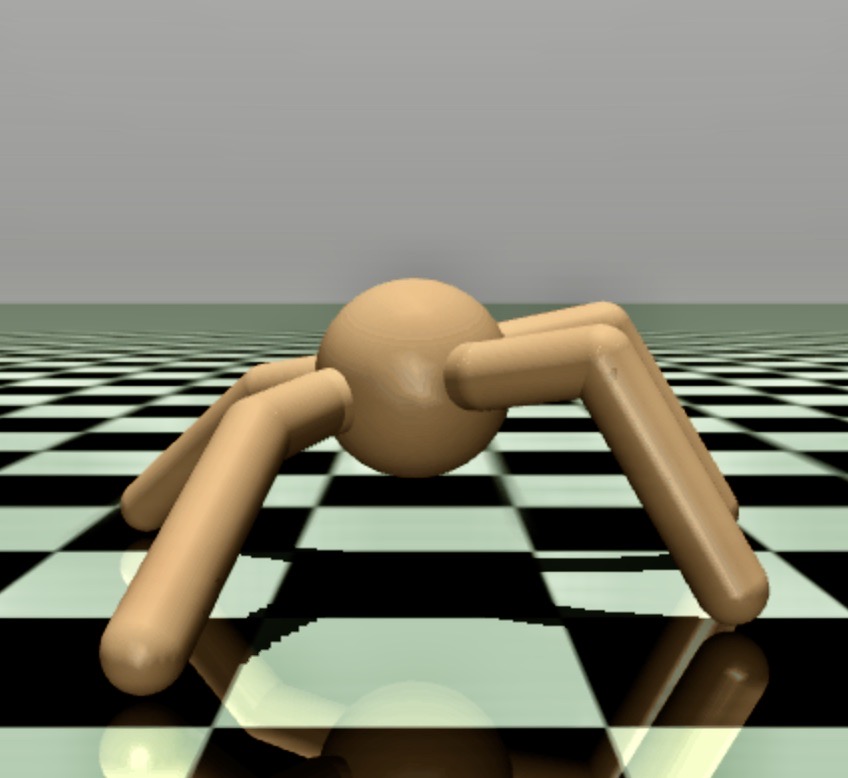}&
\includegraphics[width=2.78cm]{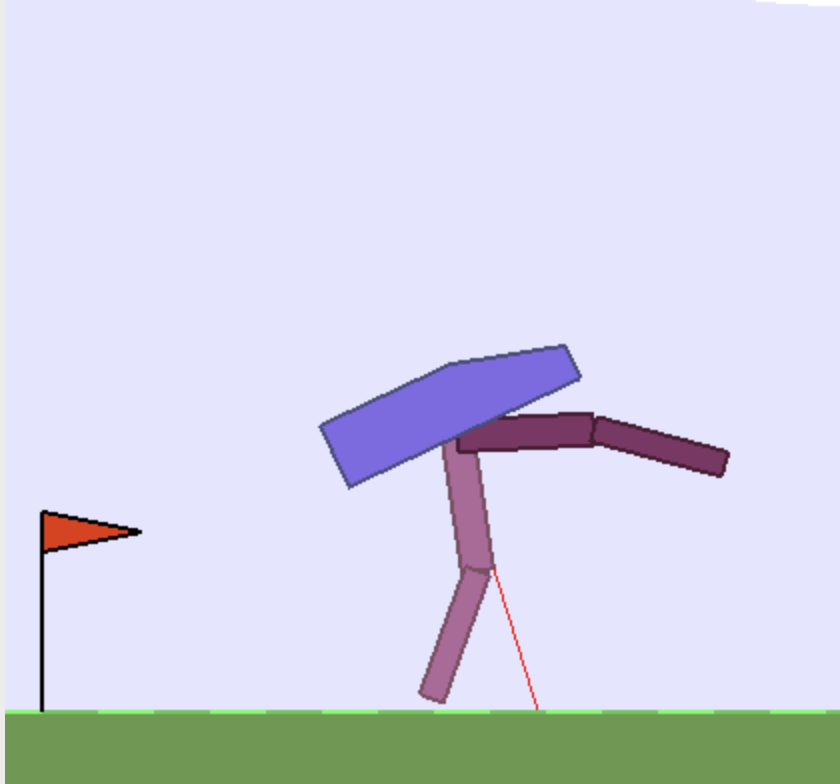}&
\includegraphics[width=2.782cm]{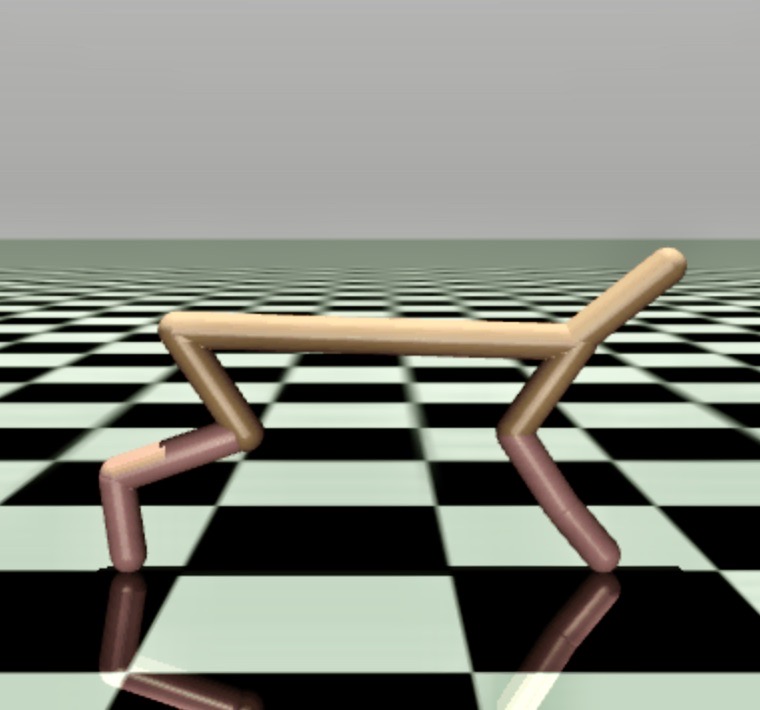} \\
%\vspace{0.5cm}
(a) Ant-v3 &
(b) {\small BipedalWalker-v3}& 
(c) {\small HalfCheetah-v3} \\
\includegraphics[width=2.8cm]{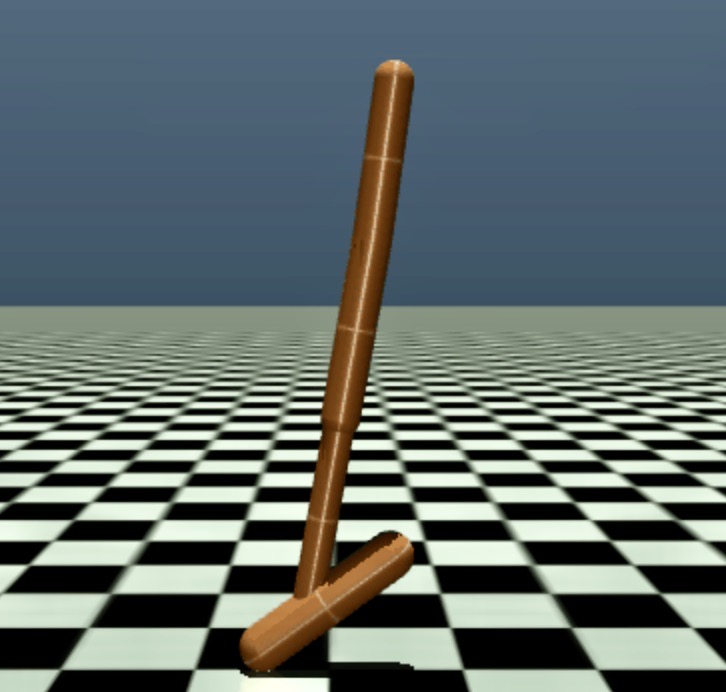}&
\includegraphics[width=2.8cm]{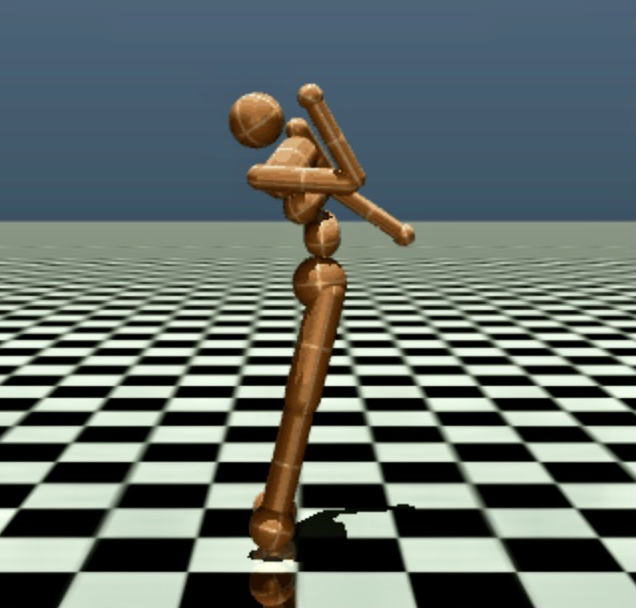}&
\includegraphics[width=2.76cm]{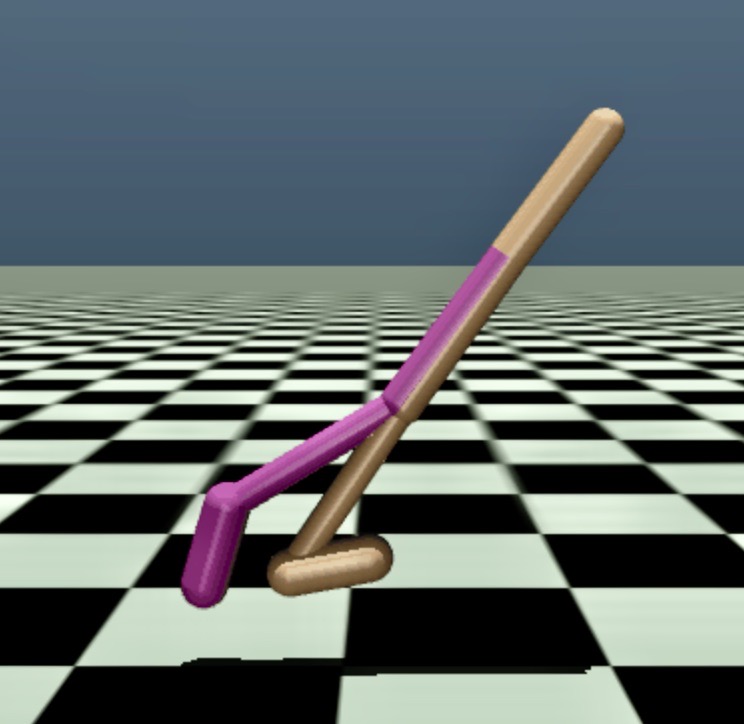}\\
(e) Hopper-v3 &
(f)  Humanoid-v3& 
(g) Walker2d-v3 \\
\end{tabular}
    \caption{Training environments (a) and (c)-(g) from MuJoCo and (b) from Box2D.}
    \vspace{-0.3cm}
    \label{fig:trainenvs} % I can do without the label too
\end{figure}

We exclusively focus on training environments  with continuous state  and action spaces.
The selection for our empirical study consists out of five MuJoCo environments \cite{Mujoco} namely: Ant-v3, Hooper-v3, HalfCheetah-v3, Humanoid-v3 and Walker2d-v3 and moreover the  BipedalWalker-v3 from Box2D \cite{catto}, see fig. \ref{fig:trainenvs}.  
\begin{figure*}[hbt!]
\begingroup
   \centering
   \renewcommand{\arraystretch}{0.2} % Default value: 1
   \tabcolsep =  -6.0pt
\begin{tabular}{ccc}
\multicolumn{3}{c}{\includegraphics[width=12.5cm]{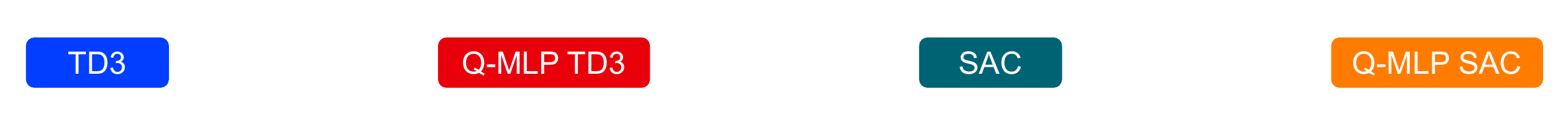}}
\vspace{-0.3cm}
 \\
\includegraphics[width=6.4cm]{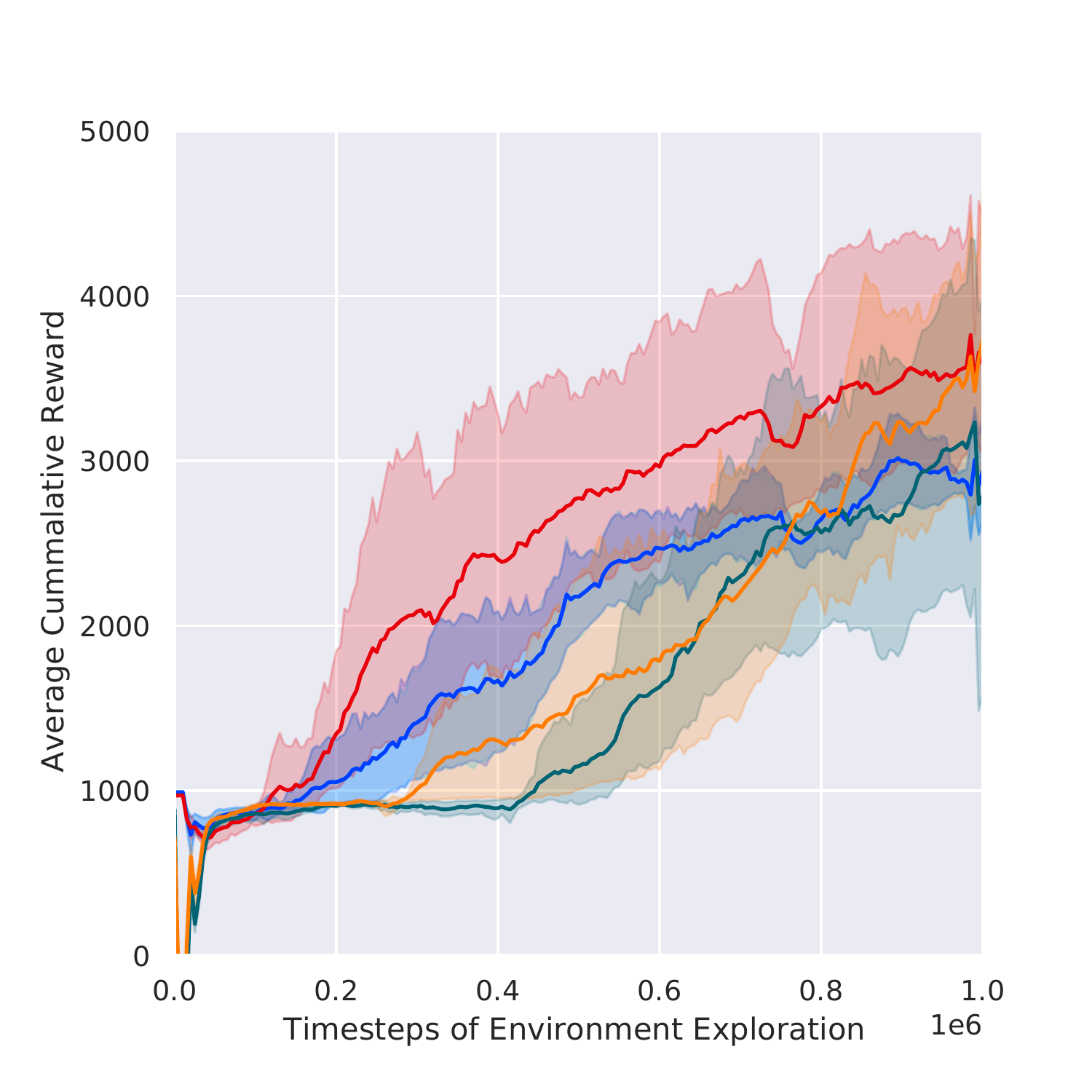}&
\includegraphics[width=6.4cm]{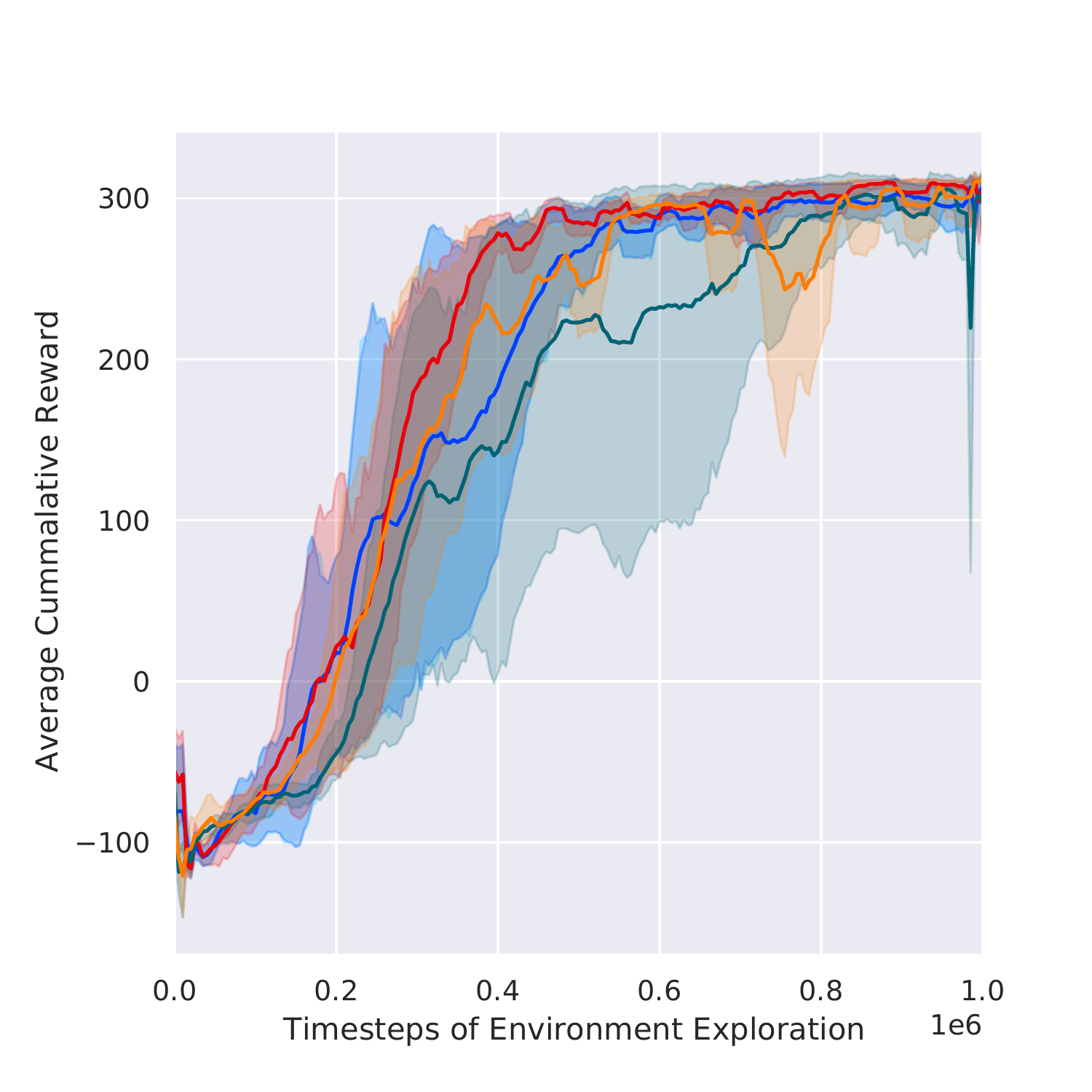}&
\includegraphics[width=6.4cm]{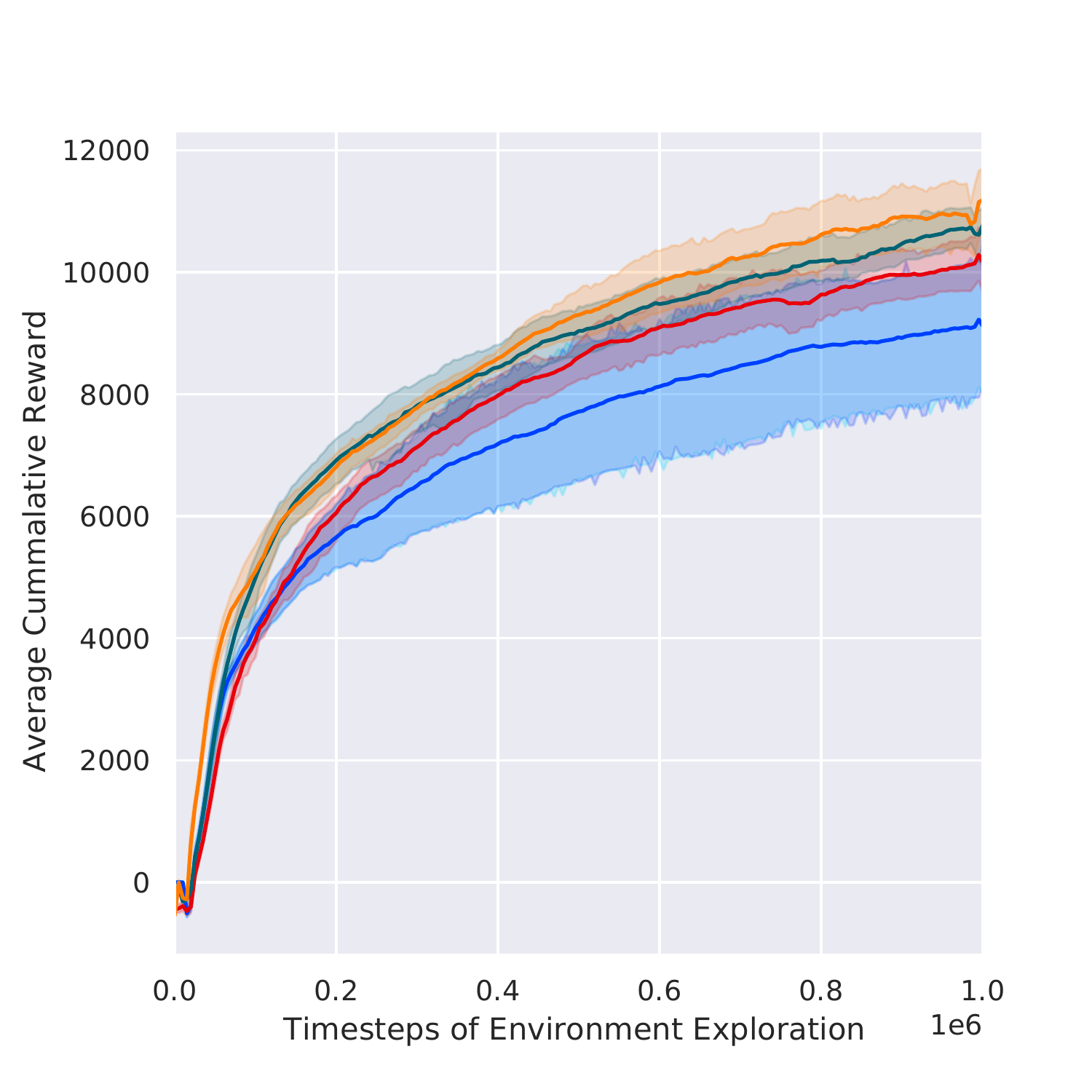} \\
%\vspace{0.5cm}
(a) Ant-v3 &
(b) BipedalWalker-v3& 
(c) HalfCheetah-v3 \\
\includegraphics[width=6.4cm]{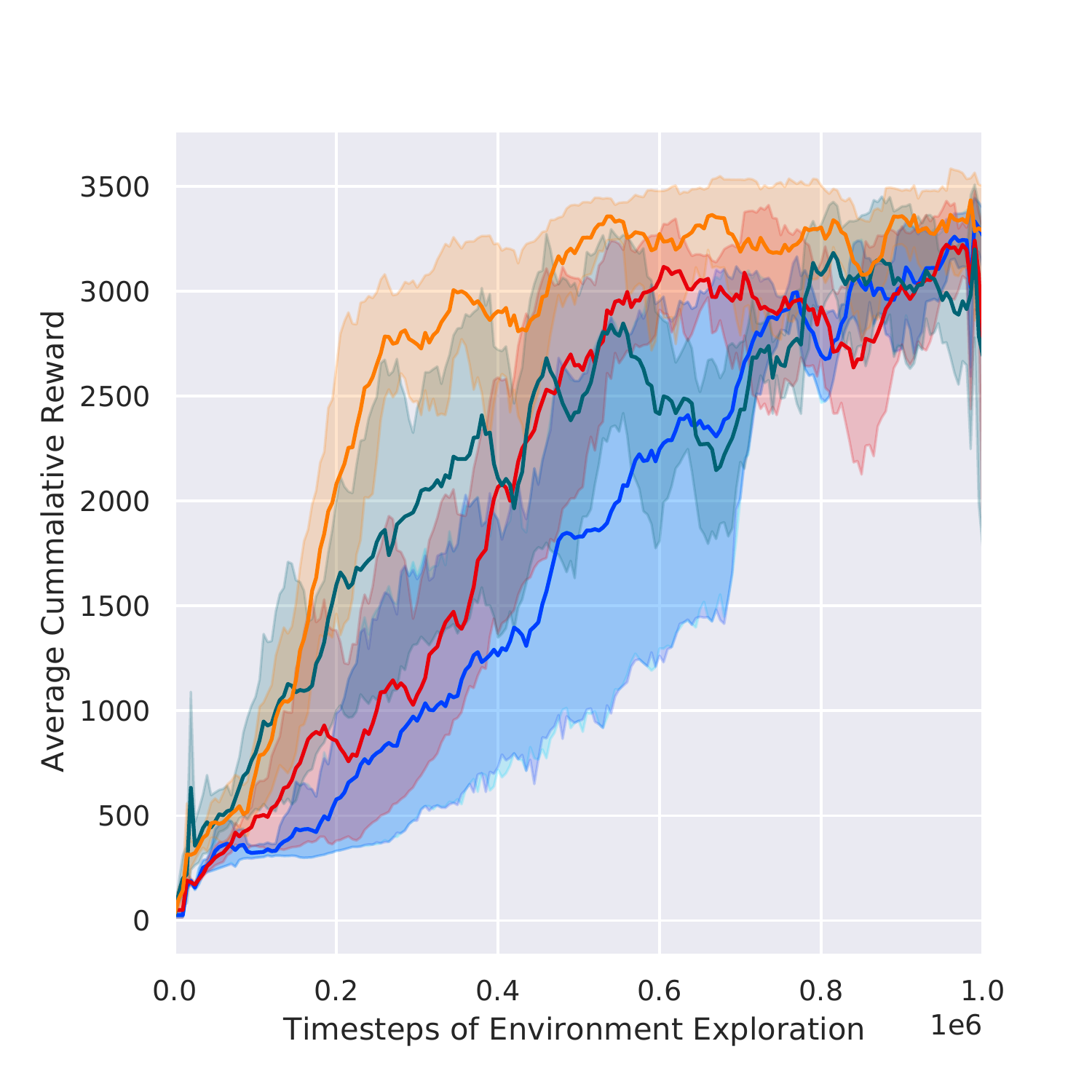}&
\includegraphics[width=6.4cm]{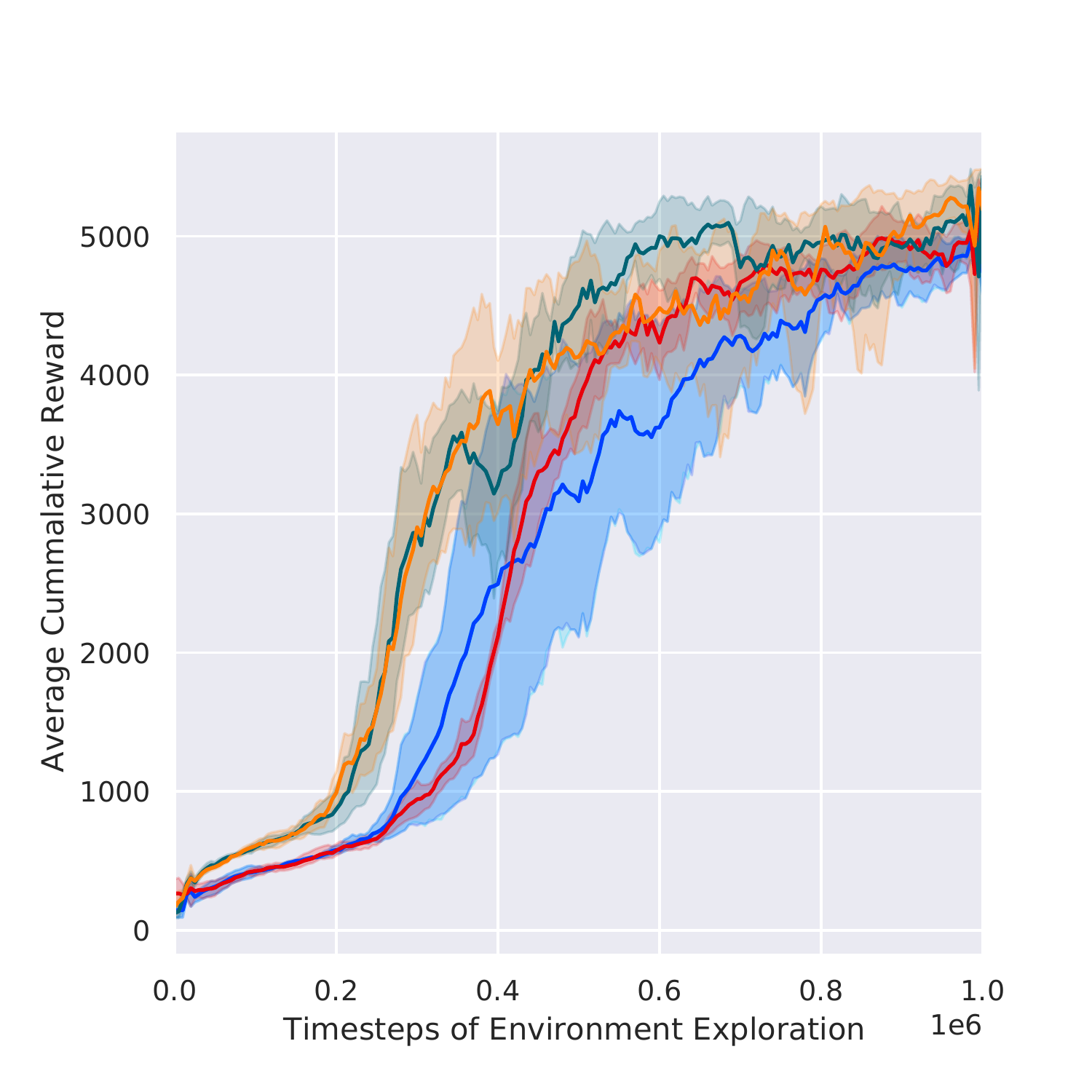}&
\includegraphics[width=6.4cm]{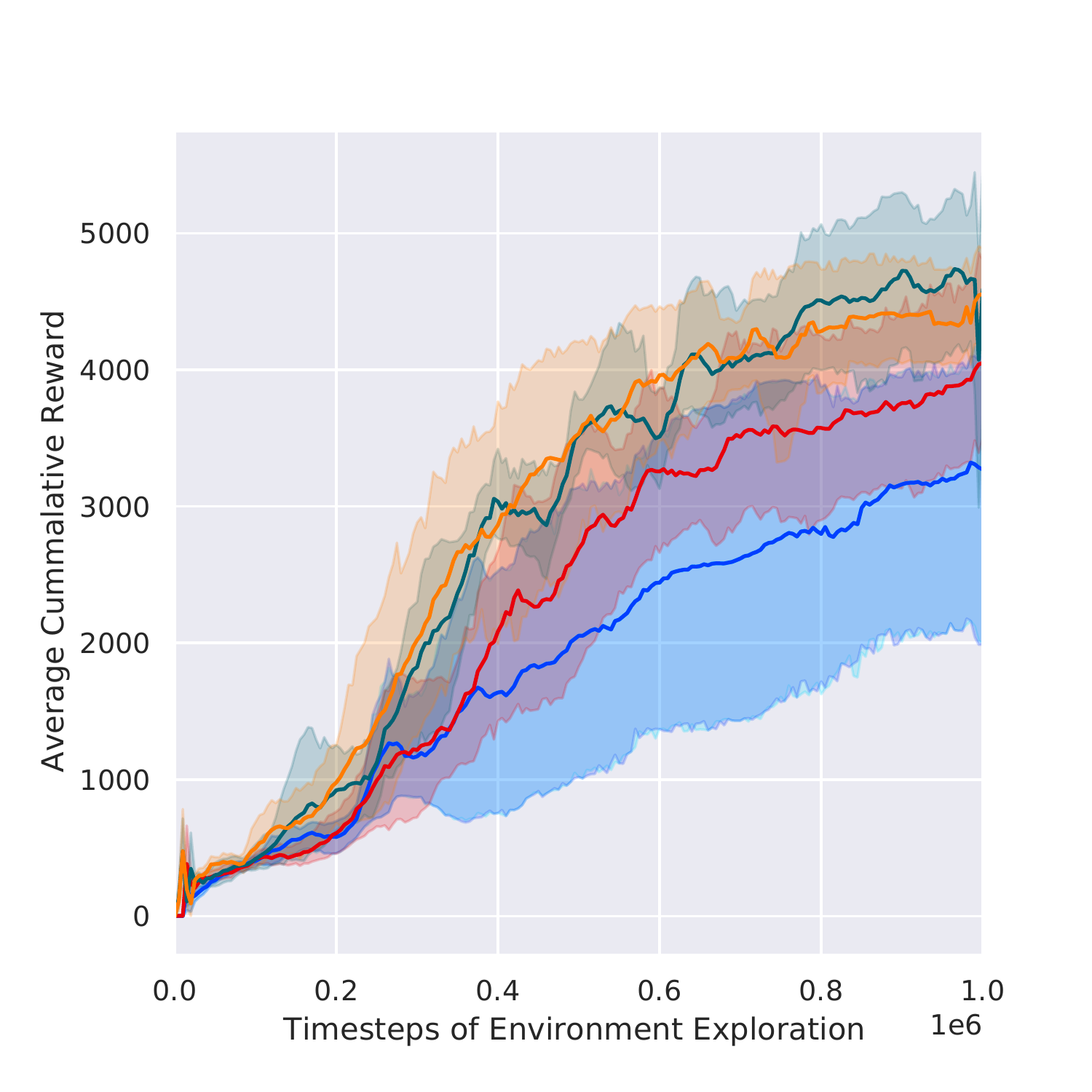}\\
(e) Hopper-v3 &
(f)  Humanoid-v3& 
(g) Walker2d-v3 \\
\end{tabular}
    \caption{Learning curves of environment (a)-(g) for both the TD3 and SAC algorithm. Mean and standard deviation averaged over five environment seeds, respectively. For better interpretability we have smoothened the plots uniformly by averaging over eight subsequent data points.}
    \label{fig:mainresults} % I can do without the label too
   \endgroup
\end{figure*}
All experiments were conducted under the same conditions. The policies are evaluated every 5k time-steps averaged over ten episodes i.e. evaluation runs with different initial conditions. We train five different seed initialisation for the random seeds $0,1,2,3,4$, respectively. The solid curves in all figures shows the average cumulative rewards per episode, and the shaded region represents the standard deviations.
The implementation of  the TD3 and SAC  algorithms as well as hyper parameter settings  are taken from  \cite{wei2021fork} \footnote{\url{https://github.com/honghaow/FORK}}. The latter is closely based on \cite{fujimoto2018addressing}\footnote{\url{https://github.com/sfujim/TD3}} in the case of the TD3 algorithm and \cite{pytorch_sac}\footnote{\url{https://github.com/ vitchyr/rlkit}  and \newline \url{https://github.com/denisyarats/pytorch_sac}} for the SAC algorithm. Thus these algorithms provide our baseline results which may be easily compared to the cited works.  %The exploration noise is added during training process with default value \cite{fujimoto2018addressing,pytorch_sac}.  

%%%%%%%%%%%%%%%%%%%%%%%%%%%%%%%%%%%%%%%%%%%%%%%%%%%%%%%%%%%%%%%%%%
\begin{figure*}[hbt!]
   \centering
   \vspace{0,2 cm}
\begin{tabular}{SSSSSSS} \toprule
    {\bf Environment} & {\bf Q$-$MLP TD3} & {\bf TD3}& {\bf Improvement} & {\bf Q$-$MLP SAC} & {\bf SAC} & {\bf Improvement} \\ \midrule
    {Ant-v3}  &  {\it 3847.40}  & 3317.29& {15.98\%} & {\bf 4027.65} &3423.88& {17.63\%}\\
    {BipedalWalker-v3}  & {\it 314.08 } & 308.48& {1.82\%} & {\bf 315.05} & { 314.46}& {0.19\%}\\
    {HalfCheetah-v3}  &  {\it 10295.54}  & 9247.73 & {11.33\%} & {\bf 11301.82} & {10884.99}& {3.83\%} \\
    {Hopper-v3} & {\it 3437.87}  & 3384.06& {1.59\%}& {\bf 3525.81}  & 3470.89 & {1.58\%} \\ \midrule
    {Humanoid-v3}  & {\it 5414.98} &5242.58 & {3.29\%} & 5494.45  &{\bf 5508.70}& {-0.26\%} \\
    {Walker2d-v3}  & {\it  4159.95} & 3485.84& {19.34\%}  & 4651.67  &  {\bf 4995.56}& {-6.88\%}  \\ \bottomrule
\end{tabular}
 \caption{The table shows the top average reward for each environment and algorithm. Bold font highlights the best return reward while italic symbolizes the winner of the comparison  of with and without quadratic neurons. The Q-MLP systematically outperforms the MLP. We separate the horizontal  line for the Humanoid as well as Walker2d environment to highlight that no improvement for the  Q-MLP SAC algorithm is found.}
\label{tab:bestRun}
\end{figure*}
\begin{figure*}[hbt!]
   \centering
\begin{tabular}{SSSSSSS} \toprule
    {\bf Environment} & {\bf Q$-$MLP TD3} & {\bf TD3}& {\bf Improvement} & {\bf Q$-$MLP SAC} & {\bf SAC} & {\bf Improvement} \\ \midrule
    {Ant-v3}  &  {\bf 575k}  & {970k}& {40.72\%} & {\it 845k} & {990k} & {14.64\%}\\
    {BipedalWalker-v3}  & {\bf 740k } & {1M} & {26.00\%} & {\it 895k} & { 970k}& {7.73\%}\\
    {HalfCheetah-v3}  &  {\bf 630k}  & {995k} & {36.68\%} & {\it 815k} & {975k}& {16.41\%} \\
    {Hopper-v3} & {\it 720k}  & {990k} & {27.27\%}& {\bf 515k}  & {795k} & {35.22\%} \\ \midrule
    {Humanoid-v3}  & {\bf 865k} &{975k} & {11.28\%} & {--}&{ --}& {$\sim$ 0\%} \\
    {Walker2d-v3}  & {\bf 565k} & {910k} & {37.91\%}  & {--}  &  {--}& {$\sim$ 0\%}  \\ \bottomrule
\end{tabular}
 \caption{ The table shows a comparison of the Q-MLP to the conventional MLP in terms of sample efficiency. Bold font highlights the best sample efficiency while italic symbolizes the winner of the comparison  of with and without quadratic neurons. For the Humanoid and Walker2d for the SAC algorithm we omit the comparison as visual inspection of figures \ref{fig:mainresults} (e) and (f) shows no quantitative difference.  }
\label{tab:sampleEfficient}
\vspace{-0,4cm}
\end{figure*}

\subsection{Q-MLP Performance on SAC \& TD3 }
\label{sec:mainResults}
In this section we discuss our empirical results.
The main results are given in fig.\,\ref{fig:mainresults} which shows  the comparison of the MLP to the Q-MLP tested with the TD3 and SAC algorithm, respectively.
For the Q-MLP  actor policy network we choose the hyper-parameters as in table \ref{tab:HypPar1}.  For the environments  with small observation space dimension  i.e. $\mathcal{O}(10)$ such as HalfCheetah-v3, BipedalWalker-v3, Hopper-v3 and Walker2d-v3 \eqref{featureFormula} can be applied simply with $\kappa = 1$. For environments with large observation space dimension $ \mathcal{O}(100)$ such as Ant-v3 and Humanoid-v3 to arrive at a less weight extensive network architecture one must choose $\kappa < 1$.
Note that for $\kappa=1$  eq.\,\eqref{featureFormula} provides the same order of degrees of freedoms as a honest quadratic neuron \eqref{QN_original}  where all the quadratic monomials are counted explicitly.
For other hyper-parameters such as details of the MLP for the Critic network see the appendix.

In summary we conclude from  fig.\,\ref{fig:mainresults} that the Q-MLP results in performance gains across  all environments and for both the TD3 and SAC algorithm. 
Let us emphasize that we do not modify the hyperparameters of the TD3 and SAC algorithm to accommodate for the Q-MLP actor policy.
In table \ref{tab:bestRun} we show the highest cumulative reward obtained averaged over the five training seeds for all environments and algorithms, respectively.
We find that even for the choices $\kappa < 1$ performance benefits are obtained, see  figure \ref{fig:mainresults}  and tables \ref{tab:bestRun},\ref{tab:sampleEfficient}.  This suggest that for those environments not all quadratic monomials  weights need to be chosen independently for the actor  to achieve high dynamic skills.
Moreover, in table \ref{tab:sampleEfficient} we provide a quantitative measure for sample efficiency of the algorithms. Namely, the time-steps it takes to achieve the best reward (table \ref{tab:bestRun}) of the weaker environment in comparison between the MLP and Q-MLP.
Let us illustrate this in an example. For the TD3 algorithm for the Ant-v3 environment the better reward is obtained by the Q-MLP, see table \ref{tab:bestRun}. Thus we compute how many time steps it takes  both the TD3 as well as the Q-MLP TD3 algorithm to achieve the "weaker" top reward, i.e. the one obtained from the MLP without quadratic neurons.
The time-steps are again averaged over the five seeds by using the mean values provided by the solid lines in the figures \ref{fig:mainresults}, respectively.
\newline
{\bf Summary of main result:}
 In conclusion, we find a benefit across almost all tasks. The top returned reward for the TD3  is  in average increased by $8.9\%$ while  being  about $29.98\%$ more sample efficient. The found improvements for the SAC algorithm are $2.7\%$  and $12.3\%$  for  top returned reward and sample efficiency, respectively. This averages to {\bf a total improvement of the cumulative reward of $5.8\%$ while being $21.1\%$ more sample efficient}.
\newline
Let us stress that the above performance measures for the Q-MLP can be improved by hyperparameter tuning. We will present details in the  section \ref{sec:ablation}. In particular, we will find that improvements can be found by increasing the number of weights of the Q-MLP in compared to the study in figure  \ref{fig:mainresults}  and tables \ref{tab:bestRun} and \ref{tab:sampleEfficient}.
Moreover, there is a clear difference on the performance gain in between the SAC and the TD3 algorithm with the latter having a strongly increased benefit of employing the Q-MLP network. As we do not perform any hyper-parameter tuning of the intrinsic values of the SAC nor the TD3 algorithm we cannot conclude on the origin of the difference.\footnote{However, it is worth noting that our implementation of the Actor Policy networks uses a Xavier uniform weight initializer which is in contrast to the Kaiming uniform initializer used for the quadratic layer.  It would be interesting to see in a future study if hyperparameter refinements such as the weights initializer for the Q-MLP SAC algorithm can lead to increased benefits across a wide range of tasks analog to the improvements found for the TD3 algorithm.}

\begin{figure*}[hpt!]
\begingroup
   \renewcommand{\arraystretch}{0.2} 
   \centering
   \tabcolsep =  - 6.0pt
\begin{tabular}{ccc}
\multicolumn{3}{c}{\includegraphics[width=12.5cm]{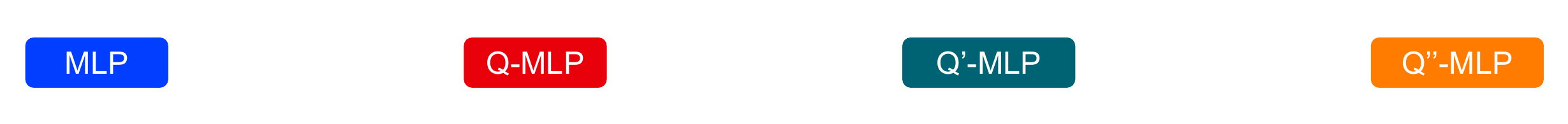}}
\vspace{-0.3cm}
 \\
\includegraphics[width=6.4cm]{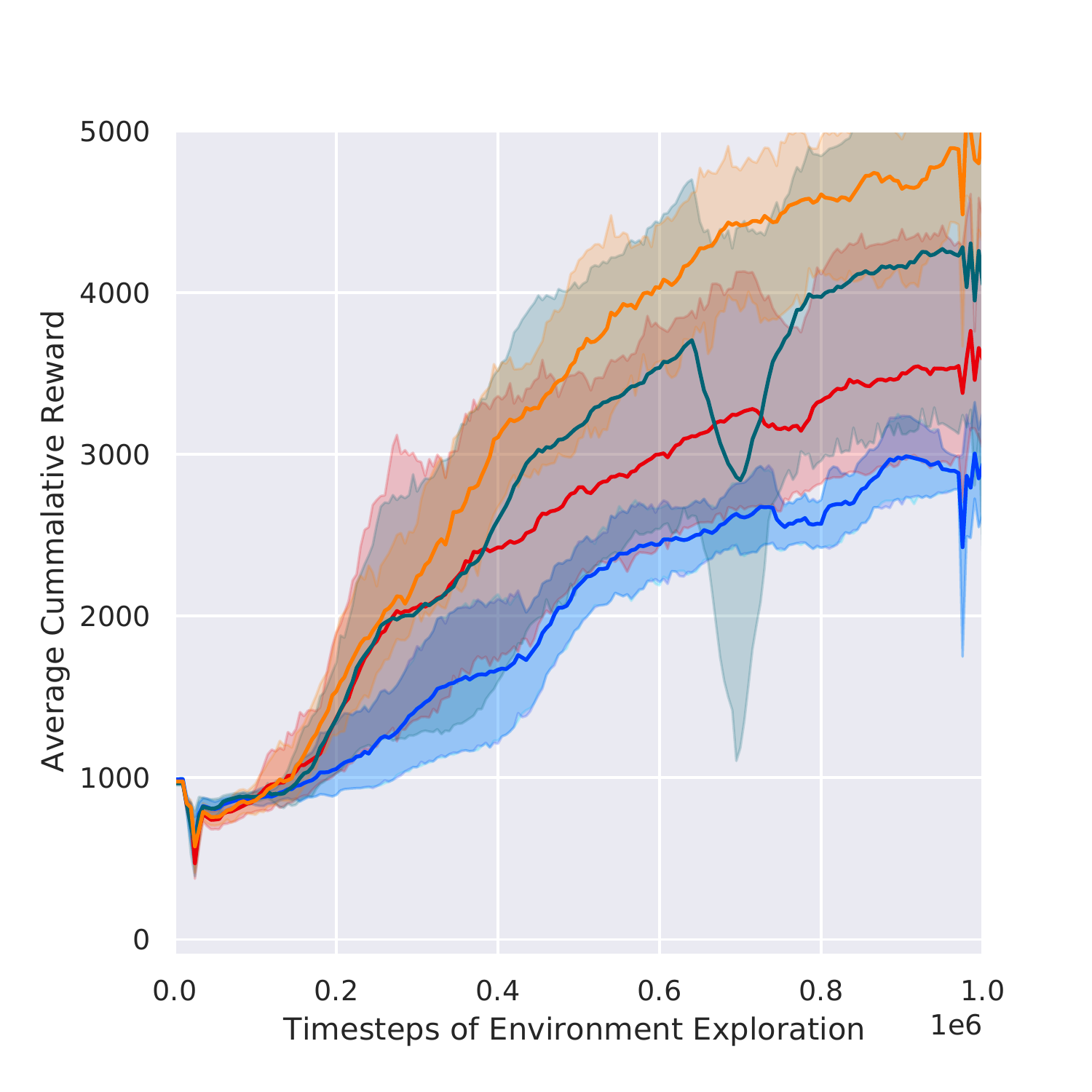}&
\includegraphics[width=6.4cm]{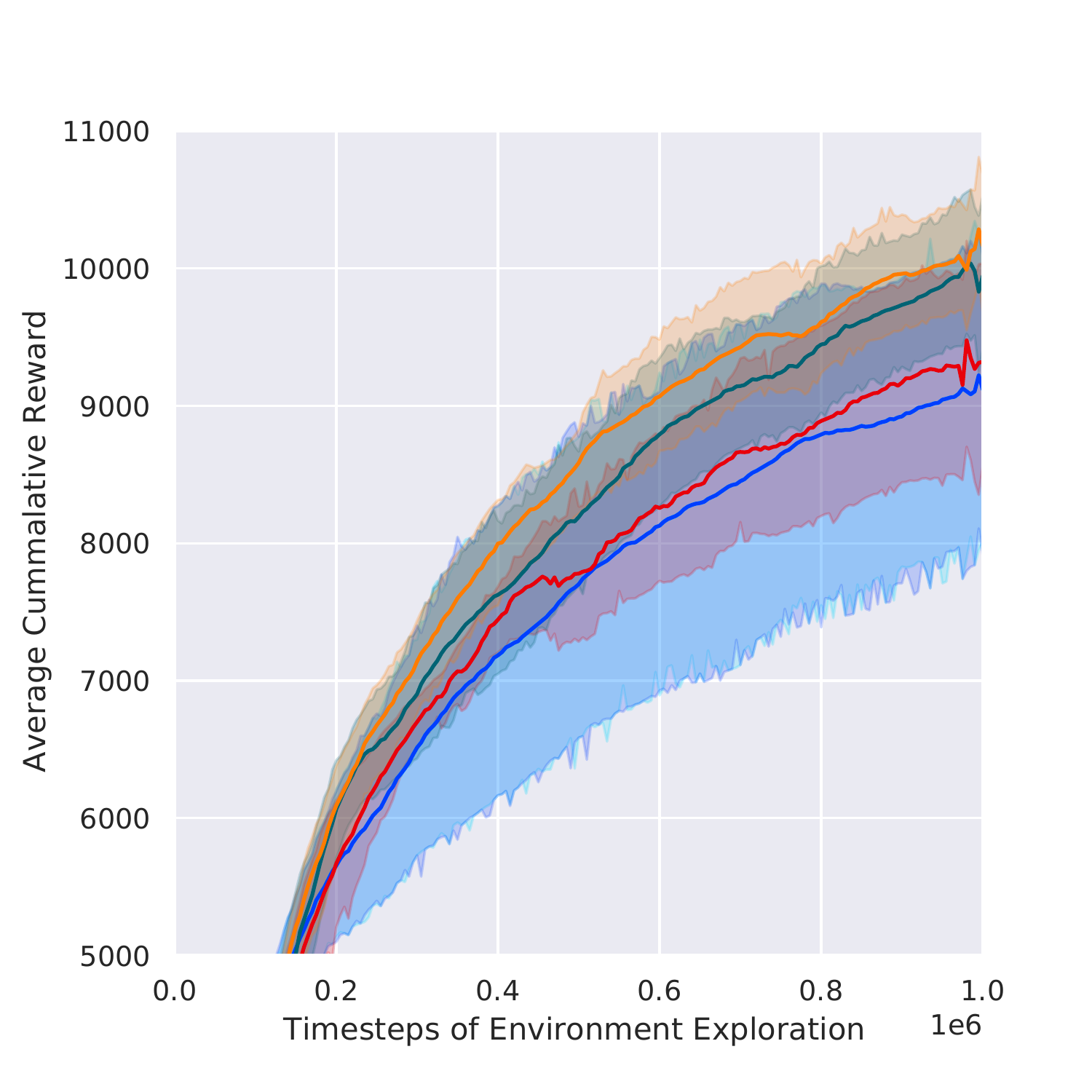}&
\includegraphics[width=6.4cm]{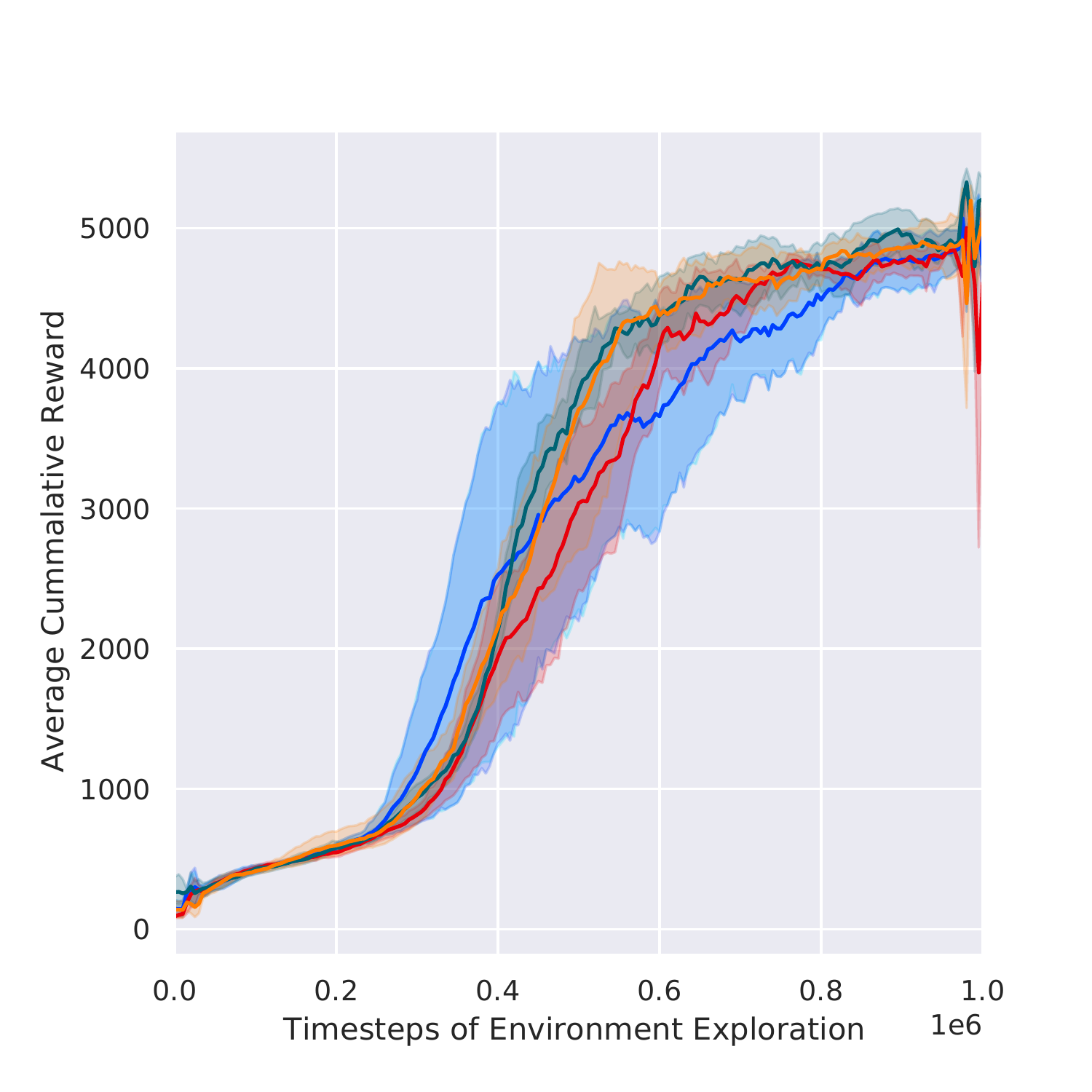}\\
%\vspace{0.5cm}
\vspace{0.2cm}
(a) Ant-v3 ;  $n_h = 64$ &
(b) HalfCheetah-v3 ; $\kappa = 1.0$ &
(c) Humanoid-v3 ;  $n_h = 192$ \\
\vspace{0.2cm}
Q, Q', Q'': $\kappa=0.1,0.25,1.0$ &
Q, Q', Q'':  $n_h = 64,128,192$ &
Q, Q', Q'': $\kappa=0.02,0.02^\ast,0.01^\ast$ \\
\end{tabular}
    \caption{Learning performance under hyperparameter changes of the Q-MLP i.e. its hidden state dimensions. The figure shows the learning curves  for the TD3 algorithm of the Ant (a), HalfCheetah (b) and Humanoid (c) environment for different hyper-parameter settings denoted by Q,Q',Q''-MLP, respectively. In (a) we vary the $\kappa$-hyper parameter i.e. the hidden features $n_f$ of the quadratic neuron while keeping $n_h$ fixed. In  (b) we keep $n_f$ fixed and vary $n_h$. Lastly, in (c) we mainly compare initializing the weights of the quadratic neuron with value zero -denoted by the asterisk   - to the default Kaiming uniform initializer.
     The precise hyper-parameter choices are found below the images in table. As for the MLP network we choose our default $n_h=256$. The plot displays the mean and standard deviation averaged over five environment seeds, respectively. For better interpretability we have smoothened the plots uniformly by averaging over 12 subsequent data points. }
    \label{fig:hypersearch} % I can do without the label too
    \endgroup
\end{figure*}
{\bf Critical comment:} While we find that in general the performance benefit holds without much hyper-parameter tuning, for the Humanoid environment however the training results may vary. As the Humanoid is by far the most complex environment this caveat may be caused due to the rich state and action space. We traced the origin to the weight initializer which is Kaiming uniform by default \cite{he2015delving}. A solution to enhance the training stability is to set the weight initializer of the quadratic neuron eq. \eqref{QNsimpledef} to zero i.e. for the weight matrices $\bf \hat\Theta' =  0$ and $\bf \hat\Theta''= 0$ as well as the bias in \eqref{QNsimpledef2}.
\newline
{\bf Noise Robustness: } We consider the TD3 algorithm for the Ant-v3 and HalfCheetah-v3 as those environments show a clear benefit in fig. \ref{fig:mainresults} and the most complex environment the Humanoid-v3 \footnote{With hyperparameters for Ant : $n_h =64 ;\, \kappa = 0.25, 0.1$, HalfCheetah: $n_h = 192 ;\, \kappa = 1.0$ and Humanoid:  $n_h = 192 ;\, \kappa = 0.02^{\ast}$.}. We find that the performance benefit of the Q-MLP is slightly improved further when adding action or observation noise at test time, see table \ref{tab:actionnoise} and \ref{tab:statenoise}.
One concludes that  for added action or observation noise the Q-MLP average  benefit of $14.8\%$  is improved to $19.7\%$ and $16.3\%$, respectively. Lastly, it is worth noting that in particular for the Humanoid environment an increase in noise leads to a higher benefit of the Q-MLP.
\begin{figure}[hbt!]
   \centering
\begin{tabular}{cccccc} \toprule
    {\bf MLP  vs. } & \multicolumn{5}{c}{\bf Action Noise Added} \\
    { \bf  Q-MLP} & {0.05} & {0.1} & {0.15} & {0.2} &{ 0.25} \\ \midrule
      {14.8\%}   &  {18.3\%}  &  { 25.0\%}  &  {24.1\%}  &  {20.2\%}  &  { 10.9\%} \\
 \bottomrule
\end{tabular}
 \caption{Q-MLP performance gain over MLP for added action noise averaged over several environments. We evaluate the model of highest no noise performance for each seed then take the average.  The displayed noise levels multiply  a random action variable. }
\label{tab:actionnoise}
   %\vspace{-0.0 cm}
\end{figure}
\begin{figure}[hbt!]
   \centering
\begin{tabular}{cccccc} \toprule
    {\bf MLP vs. } & \multicolumn{5}{c}{\bf Observation Noise Added} \\
     { \bf Q-MLP}& {0.01} & {0.02} & {0.03} & {0.04} &{ 0.05} \\ \midrule
     {14.8\%}  &  {13.7\%}  &  { 16.3\%}  &  {18.7\%}  &  {17.6\%}  &  {15.0\%} \\
    \bottomrule
\end{tabular}
 \caption{Q-MLP performance gain over MLP for added observation noise averaged over several environments. We evaluate the model of highest no noise performance for each seed then take the average. The displayed noise levels multiply a random observation variable. }
\label{tab:statenoise}
   \vspace{-0.2 cm}
\end{figure}

\newpage
\subsection{Ablation Study}
\label{sec:ablation}
\begin{figure*}[hpt!]
\begingroup
   \renewcommand{\arraystretch}{0.2} 
   \centering
 \tabcolsep =  -6.0pt
\begin{tabular}{ccc}
\multicolumn{3}{c}{\includegraphics[width=12.5cm]{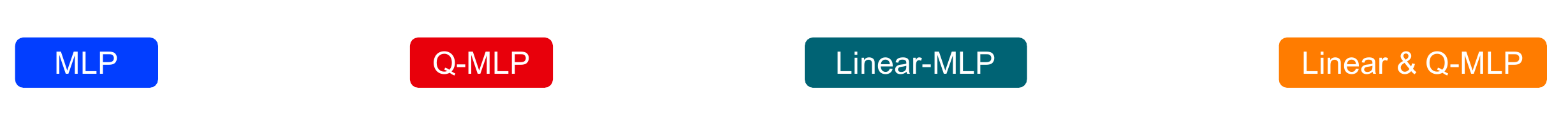}}
\vspace{-0.3cm}
 \\
\includegraphics[width=6.4cm]{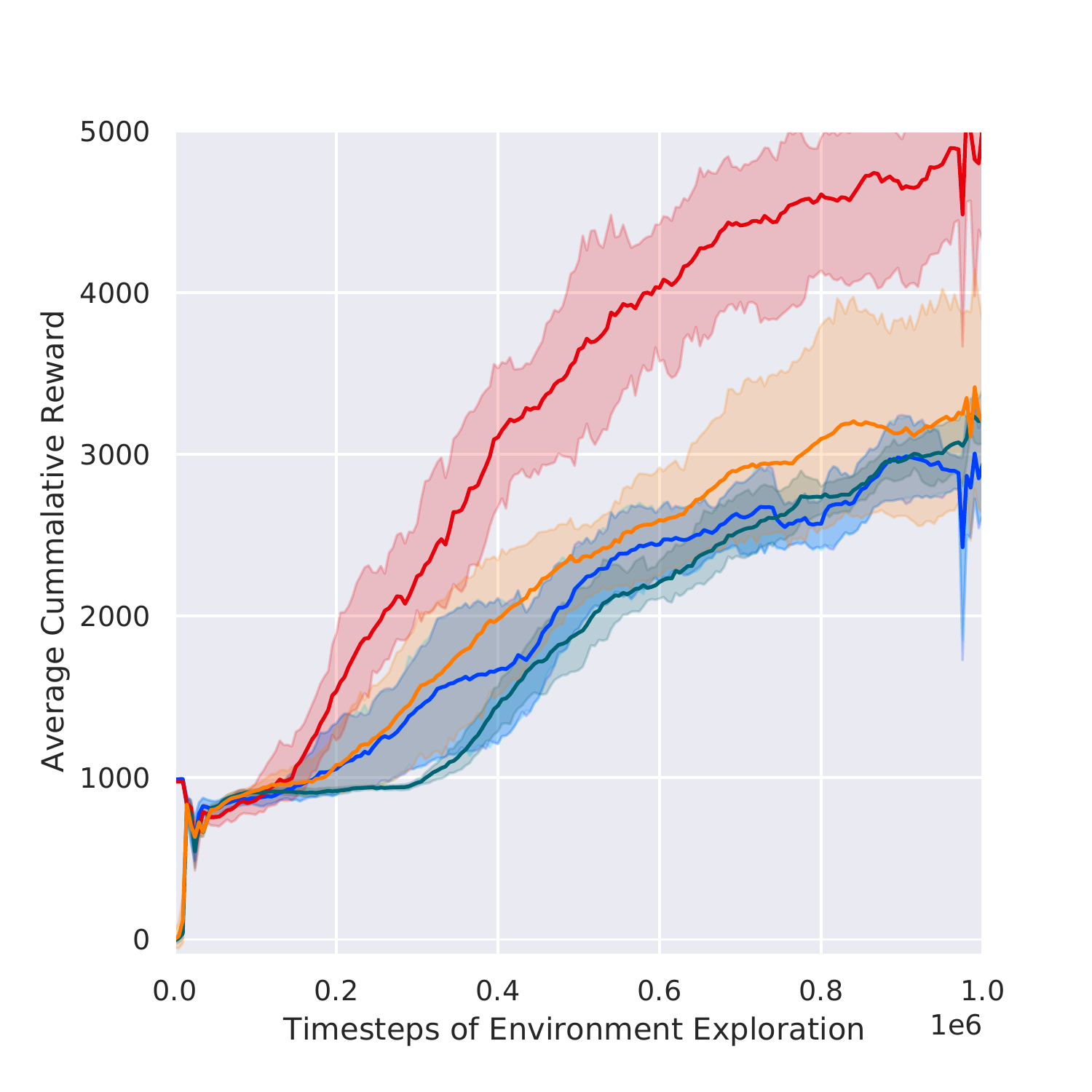}&
\includegraphics[width=6.4cm]{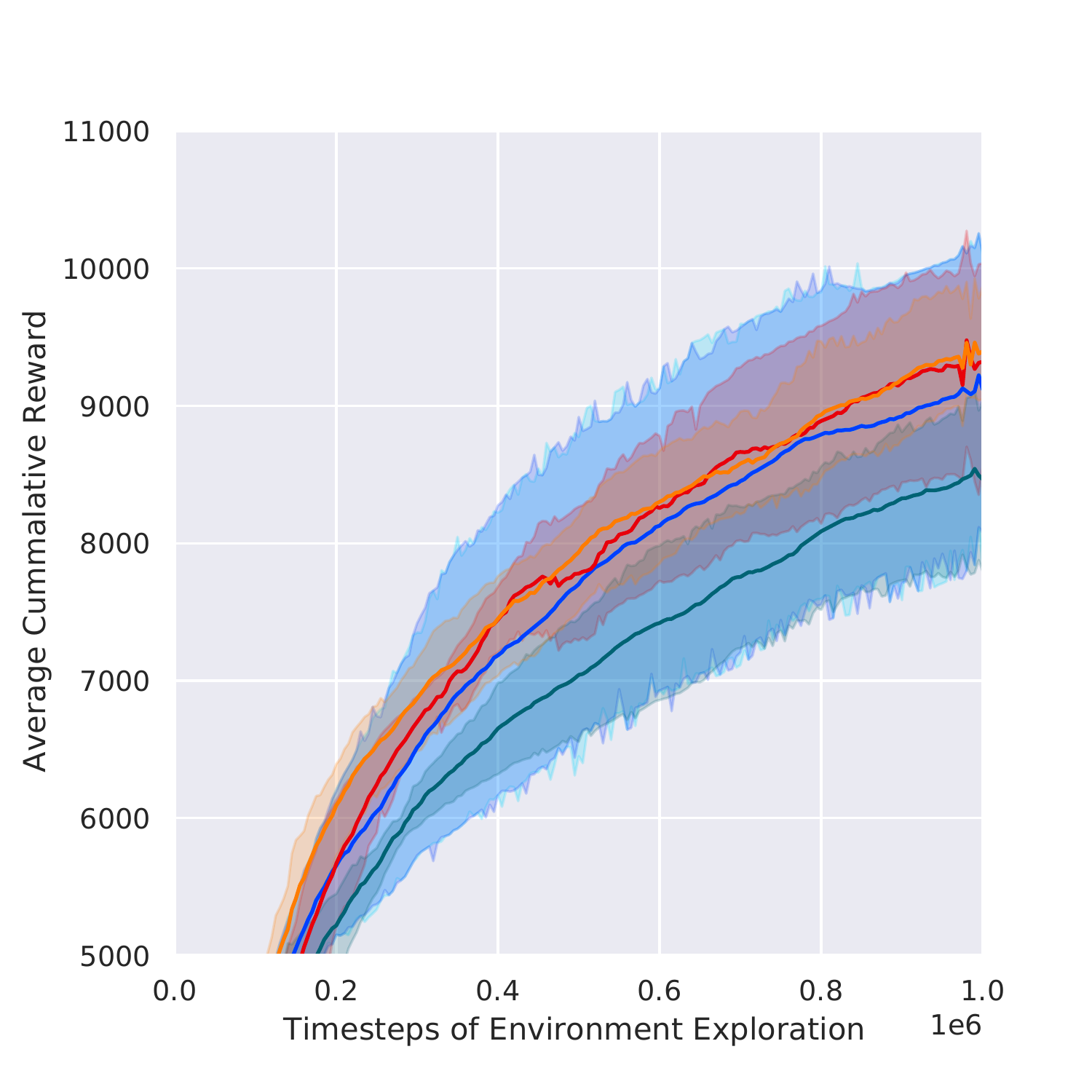}&%Ablation_BipedalWalker-v3.pdf}&
\includegraphics[width=6.4cm]{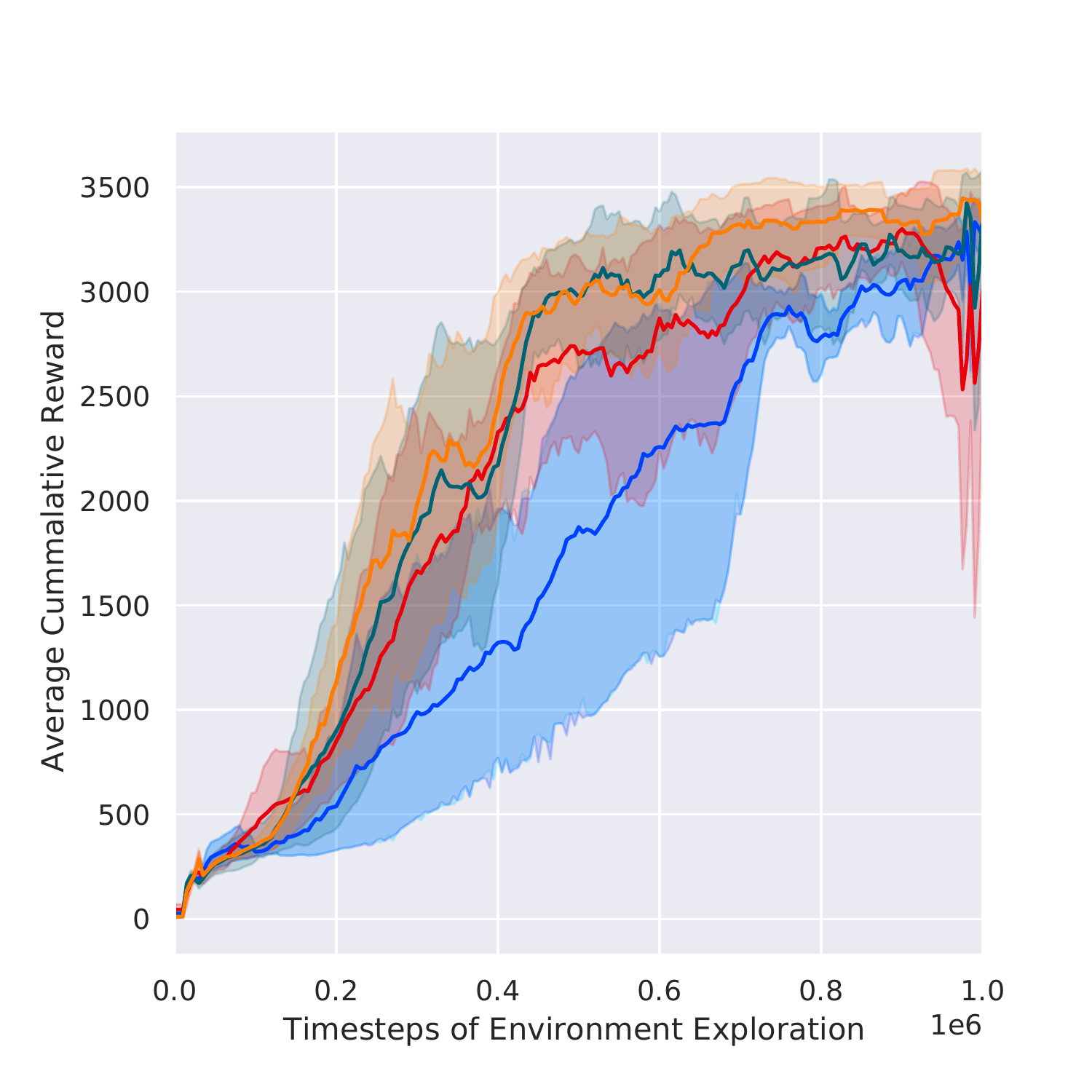} \\
%\vspace{0.5cm}
(a) Ant-v3 &
(b) HalfCheetah-v3& 
(c)  Hopper-v3  \\
%\includegraphics[width=6.0cm]{figures/Ablation_Hopper-v3.pdf}&
%\includegraphics[width=6.0cm]{figures/Ablation_Ant-v3.pdf}&%Ablation_Humanoid-v3.pdf}&
%\includegraphics[width=6.0cm]{figures/Ablation_Ant-v3.pdf}\\%Ablation_Walker2d-v3.pdf}\\
%(e) Hopper-v3 &
%(f)  Humanoid-v3& 
%(g) Walker2d-v3 \\
\end{tabular}
    \caption{Ablation study in regard to the network alteration containing a linear unit, see fig.'s \ref{L-MLP}  and \ref{LQ-MLP}. The plot shows the learning curves  mean and standard deviation derived from five environment seeds, respectively. For better interpretability we have smoothened the plots uniformly by averaging over 12 subsequent data points.}
    \label{fig:ablationresults} 
    \endgroup
     \vspace{-0.3 cm}
\end{figure*}
In this section we  perform an ablation study to compare the Q-MLP network to the alteration containing linear units see fig.'s\,\ref{L-MLP}  and \ref{LQ-MLP}. 
As well as discuss quantitative features of changes in the performance of the Q-MLP in regard to changes of the hyper-parameters $n_h$ and $\kappa$ i.e.\,$n_f$.
For simplicity we restrict this study to only the TD3 algorithm in this section and to a reduced number of environments.

Let us start our  study by discussing systematics of our approach in regard to changes of the hyper-parameters $n_f$ and $n_h$ as well as the weight initializers for the quadratic neuron layers, see figure \ref{fig:hypersearch}.
\newline
{\bf Fixed $n_h$ and varying $n_f \,\& \,\kappa$:} In figure \ref{fig:hypersearch} (a) we display the learning curves for the Ant-v3 environment for increasing $\kappa$ i.e. $n_f$ and find that there is a direct positive  correlation  to the reward returned as well as the sample efficiency.
In particular, we find that the benefit over the conventional MLP doubles from $\sim 16 \%$ to $\sim 33 \%$ for the top reward and from $\sim 40\%$ to $\sim 60\%$ for the sample efficiency when increasing the number of weights of the quadratic layer compared to the one used in table \ref{tab:bestRun}.
\newline
{\bf Fixed $n_f \,\& \,\kappa$ and varying $n_h$ :} In figure \ref{fig:hypersearch} (b) we display the learning curves for the HalfCheetah-v3 environment for increasing $n_h$ and analogously find that there is a direct positive correlation  to the reward returned as well as the sample efficiency.
\newline
{\bf Initializer ablation study:}  In figure \ref{fig:hypersearch} (c) we display the learning curves for the Humanoid-v3 environment. The focus is on the stability achieved by initializing the quadratic neuron weights with zero. For details see our comments earlier in the text. From this study we see that initializing the weights with zero not only leads to improved training stability over different hyperparameter choices. Furthermore, it  slightly improves the sample efficiency.

Let us next turn our attention to the comparison of the Q-MLP network to the alteration containing linear units see figures \ref{L-MLP}  and \ref{LQ-MLP}. For all experiments we choose $n_h=64$ and $\kappa = 1.0$.
The results can be found in figure \ref{fig:ablationresults}. 
\newline
{\bf Main conclusion on linear units:} Our results figure \ref{fig:ablationresults} suggest that  the benefit of using the linear unit is negligible and more likely leads to worse performance. 
Across the tasks (a) and (b) from  fig.\,\ref{fig:ablationresults} one concludes that using the linear unit alone leads to a drop in the returned reward. In the HalfCheetah environment fig.\,\ref{fig:ablationresults}(b)  the network alteration with both a quadratic and linear neuron leads to a minor benefit over the Q-MLP alone.
Whereas in the experiment fig.\,\ref{fig:ablationresults}(c) the latter leads to a visible performance increase. 
In conclusion, for both alterations the downsides exceed the benefit. Moreover, let us  emphasise that our results here do not reproduce the performance gains found in the original study of linear units \cite{srouji2018structured}.
It is our opinion that this is due to the fact that our TD3 and SAC algorithm  produce significantly higher rewards as the ones in  \cite{srouji2018structured}. This is a critical point which we have observed using less sophisticated reinforcement learning algorithms: benefits due to  changes in the actor policy which occur when using a weak RL algorithm often fade away when employing a state of the art algorithm. In other words, it is a much bigger challenge to have architectural modifications of the actor policy improve the performance of an already strong learning setup - such as the one used in this work -.

%%%%%%%%%%%%%%%%%%%%%%%%%%%%%%%%%%%%%%%%%%%%%%%%%%%%%%%%%%%%%%%%%%
\vspace{0,3 cm}
\section{CONCLUSIONS}
\vspace{0,1 cm}
In this work we propose Q-MLP, a modification of the conventional MLP actor policy network to be used in Actor-Critic algorithms. Our contribution is the additional direct quadratic neuron connection between observation state input and action state output.  Our empirical results derived by employing two state-of-the-art model-free reinforcement learning algorithms across six environments show performance improvements in terms of returned reward and sample efficiency. All while maintaining a smaller number of weights compared to the baseline provided by a conventional MLP actor policy network.

\vspace{0,3 cm}
\section*{APPENDIX}
\vspace{0,1 cm}
The neural networks are trained using Pytorch $1.7.1+\text{cu}110$ and Python $3.8.7$.
The actor policy MLP as well as the the MLP part of the Q-MLP have two hidden layers with  ReLU activation and hidden dimension units given in table \ref{tab:HypPar1}, respectively. The final activation function  of the Q-MLP is $tanh$. For the critic network for both the TD3 as well as the SAC algorithm we use a MLP with two hidden layers with $256$ units each and ReLU non-linearity. Moreover, the optimizer is Adam with learning rate $3 \cdot 10^{-4}$, discount factor $\gamma = 0.99$, replay buffer size $1M$, target update rate of $5\cdot 10^{-3}$ and a batch size of $100$ for both the TD3 as well as SAC. For the TD3 we us a target update delay of $2$, a policy noise of $0.2$ with a noise clip of $0.5$.
\begin{figure*}[hbt!]
   \centering
    \vspace{0,2 cm}
\begin{tabular}{ccccccccccccc} \toprule
    {} & \multicolumn{4}{c}{{\bf Q$-$MLP TD3}} & \multicolumn{2}{c}{{\bf TD3}} &  \multicolumn{4}{c}{{\bf Q$-$MLP SAC}} & \multicolumn{2}{c}{{\bf SAC}}  \\ \midrule
    {\bf Environment} & {\bf $n_h$} & {$\kappa$} & {\bf $n_f$}& {\bf weights} & {\bf $n_h$} &  {\bf weights} &{\bf $n_h$} &{$\kappa$}& {\bf $n_f$}& {\bf weights} &{\bf $n_h$} & {\bf weights}  \\ \midrule
    {Ant-v3}  & 64 & 0.1 & 62 & {\bf 26.1k} & 256 & {96.5k} & 128& 0.1 & 62& {\it 47.7k} &256 & {98.6k} \\
    {BipedalWalker-v3}   & 192 & 1.0 & 300 & {\bf 58.2k} & 256 & {73.2k} & 192 &1.0 & 300& {\it 60.2k} &256 & {74.2k}   \\
    {HalfCheetah-v3}  & 192 & 1.0 & 153 & {\bf 47.8k} & 256 & {71.9k} & 192 &1.0 & 153& {\it 49.9k} &256 & {73.5k}  \\
    {Hopper-v3}   & 128 & 1.0 & 66 & {\bf 20.1k} & 256 & {69.6k} & 128& 1.0 & 66& {\it 20.6k} &256 & {70.4k} \\ \midrule
    {Humanoid-v3}  & 192 & 0.02& {28$^\ast$} & {\bf 134.2k} & 256 & {166.7k} & 224 &  0.02 & {28$^\ast$}& {\it 164.5k} &256 & {171.0k}  \\
    {Walker2d-v3}   & 64 & 1.0& 153 & {\bf 11.8k} & 256 & {71.9k} & {128} & 1.0&  153& {\it 27.4k} &256 & {73.5k}   \\ \bottomrule
\end{tabular}
 \caption{ Hyperparameter details of the actor policy networks. Hidden dimensions and total weight numbers of the Q-MLP and MLP neural networks used in the the main study see figures \ref{fig:mainresults} and tables \ref{tab:bestRun} and \ref{tab:sampleEfficient}. Note that the total number of weights is influenced by the observation and action  space dimensions of the environment. Moreover, as the SAC algorithm admits a deterministic as well as stochastic policy each admitting the same number of weights displayed in the table, respectively. The asterisk  symbolises that we have initialized the quadratic neuron with zero weights in contrast to the standard Kaiming uniform initializer \cite{he2015delving} used in pytorch \cite{NEURIPS2019_9015}.}
 \vspace{-0.6 cm}
\label{tab:HypPar1}
\end{figure*}

%\begin{figure*}[hbt!]
 %  \centering
%\begin{tabular}{cccccccccccc} \toprule
  %  {} & \multicolumn{3}{c}{{\bf Q$-$MLP TD3}} & \multicolumn{2}{c}{{\bf TD3}} & \multicolumn{3}{c}{{\bf Q'$-$MLP TD3}} & \multicolumn{3}{c}{{\bf Q''$-$MLP TD3}} \\ \midrule
  %  {\bf Environment} & {\bf $n_h$} & {\bf $n_f$}& {\bf weights} & {\bf $n_h$} &  {\bf weights} &{\bf $n_h$} & {\bf $n_f$}& {\bf weights} &{\bf $n_h$}&{\bf $n_f$} & {\bf weights}  \\ \midrule
   % {Ant-v3}  & 64 & 20 & {\bf 10k} & 256 & {90k} & 64 & 20& {\it 10k} &256& 20 & {90k} \\
  %  {HalfCheetah-v3}  & 64 & 20 & {\bf 10k} & 256 & {90k} & 64 & 20& {\it 10k} &256 & 20& {90k}  \\
  %  {Humanoid-v3}   & 64 & 20 & {\bf 10k} & 256 & {90k} & 64 & 20& {\it 10k} &256& 20 & {90k} \\ \bottomrule
%\end{tabular}
% \caption{\MWnote{Values filled in soon...} Hidden dimensions and total weight numbers of the Q-MLP and MLP neural networks used in the the main study see figures \ref{fig:mainresults} and tables \ref{tab:bestRun} and \ref{tab:sampleEfficient}. Note that the total number of weights is influenced by the observation and actions a space dimensions of the environment }
%\label{tab:HypPar2}
%\end{figure*}

%\MWnote{ACKNOWLEDGMENT to appear in acceptd version only?}
\vspace{0,5 cm}
\section*{ACKNOWLEDGMENT}
\vspace{0,2 cm}
We would like to thank Yoshiteru Nishimura for his technical assistance.
The current work is, partly, supported by JSPS KAKENHI Grant Number JP18H03287 and JST CREST Grant Number JPMJCR1913.

%\MWnote{The comment  "addtolength"   is relevant.}
%\addtolength{\textheight}{-1cm}   % This command serves to balance the column lengths

%%%%%%%%%%%%%%%%%%%%%%%%%%%%%%%%%%%%%%%%%%%%%%%%%%%%%%%%%%%%%%%%%%%%%%%%%%%%%%%%

%%%%%%%%%%%%%%%%%%%%%%%%%%%%%%%%%%%%%%%%%%%%%%%%%%%%%%%%%%%%%%%%%%%%%%%%%%%%%%%%

%%%%%%%%%%%%%%%%%%%%%%%%%%%%%%%%%%%%%%%%%%%%%%%%%%%%%%%%%%%%%%%%%%%%%%%%%%%%%%%%
%\section*{APPENDIX}

%\section*{ACKNOWLEDGMENT}

%%%%%%%%%%%%%%%%%%%%%%%%%%%%%%%%%%%%%%%%%%%%%%%%%%%%%%%%%%%%%%%%%%%%%%%%%%%%%%%%
%\newpage
\vspace{0,5 cm}
\nocite{*}
\bibliographystyle{IEEEtran} 
%\newpage
\bibliography{references}

\end{document}